\documentclass[conference]{IEEEtran}
\IEEEoverridecommandlockouts

\usepackage{graphicx}
\usepackage{caption}
\usepackage{amsmath,amssymb,enumerate}

\usepackage{amsthm}
\usepackage{mathtools}
\usepackage{breqn}
\usepackage[usenames,dvipsnames,svgnames,table, xcdraw]{xcolor}
\usepackage[colorlinks=true,pdfpagemode=UseNone,citecolor=black,linkcolor=black,urlcolor=BrickRed]{hyperref}
\usepackage{algorithm, algpseudocode}

\usepackage{blindtext}
\usepackage{gensymb}
\usepackage{xparse}
\usepackage{lipsum}
\usepackage{mathrsfs}
\usepackage[mathscr]{euscript}
\usepackage{times}
\usepackage[noadjust]{cite} 
\usepackage{multicol}
\usepackage[caption=false,font=normalsize]{subfig}
\usepackage{amsfonts}
\usepackage[utf8]{inputenc}
\usepackage[T1]{fontenc}
\usepackage{textcomp}
\usepackage{balance}
\usepackage{amsfonts}
\usepackage{soul}
\usepackage{multirow}
\usepackage{booktabs}
\usepackage{sidecap}
\usepackage{makecell}
\usepackage{array,float}
\usepackage{marginnote}

\renewcommand{\thefigure}{\arabic{figure}}

\definecolor{Orange}{rgb}{0.9608, 0.5882, 0.3922}
\definecolor{Yellow}{rgb}{0.9608, 0.9020, 0.3922}
\definecolor{Brown}{rgb}{0.5882, 0.2353, 0.1176}
\definecolor{Maroon}{rgb}{0.7059, 0.1176, 0.3137}
\definecolor{Grapefruit}{rgb}{1, 0.3137, 0.3922}
\definecolor{Royal Blue}{rgb}{0.1176, 0.1176, 1}
\definecolor{Magenta}{rgb}{0.7843, 0.1569, 1}
\definecolor{Pink}{rgb}{1, 0.5882, 1}
\definecolor{Purple}{rgb}{0.2941, 0, 0.2941}
\definecolor{White}{rgb}{1, 1, 1}
\definecolor{Black}{rgb}{0, 0, 0}
\definecolor{Light Green}{rgb}{0.3411, 0.9764, 0.4274}
\definecolor{Light Red}{rgb}{0.9764, 0.3411, 0.3411}
\definecolor{Yellow}{rgb}{1, 0.9294, 0.4784}

\definecolor{unknownColor}{RGB}{0, 0, 0}
\definecolor{concreteColor}{RGB}{97, 171, 47}
\definecolor{grassColor}{RGB}{159, 77, 200}
\definecolor{rocksColor}{RGB}{141, 49, 126}
\definecolor{speedwayBricksColor}{RGB}{235, 128, 55}
\definecolor{redBricksColor}{RGB}{174, 149, 8}
\definecolor{pebblePavementColor}{RGB}{98, 3, 141}
\definecolor{lightMarbleTilingColor}{RGB}{74, 110, 203}
\definecolor{darkMarbleTilingColor}{RGB}{115, 240, 49}
\definecolor{dirtPathsColor}{RGB}{127, 57, 78}
\definecolor{roadPavementColor}{RGB}{142, 143, 60}
\definecolor{shortVegetationColor}{RGB}{17, 187, 187}
\definecolor{porcelainTileColor}{RGB}{165, 247, 137}
\definecolor{metalGratesColor}{RGB}{27, 183, 89}
\definecolor{blondMarbleTilingColor}{RGB}{80, 29, 134}
\definecolor{woodPanelColor}{RGB}{244, 81, 150}
\definecolor{patternedTileColor}{RGB}{159, 77, 163}
\definecolor{carpetColor}{RGB}{116, 100, 60}
\definecolor{crosswalkColor}{RGB}{153, 207, 156}
\definecolor{domeMatColor}{RGB}{159, 138, 135}
\definecolor{stairsColor}{RGB}{131, 217, 44}
\definecolor{doorMatColor}{RGB}{131, 97, 123}
\definecolor{thresholdColor}{RGB}{101, 226, 115}
\definecolor{metalFloorColor}{RGB}{40, 43, 156}
\definecolor{unlabeledColor}{RGB}{0, 0, 0}

\newcommand{\besttred}[1]{\textcolor{red}{\textbf{#1}}}
\newcommand{\bestbev}[1]{\textcolor{blue}{\textbf{#1}}}

\newcommand{\coda}{CODa}

\newcommand{\seclabel}[1]{\label{sec:#1}}
\newcommand{\secref}[1]{Section~\ref{sec:#1}}

\newcommand{\figlabel}[1]{\label{fig:#1}}
\newcommand{\figref}[1]{Fig.~\ref{fig:#1}}
\newcommand{\tablabel}[1]{\label{tab:#1}}
\newcommand{\tabref}[1]{Table~\ref{tab:#1}}

\usepackage{xcolor}
\usepackage{import}
\usepackage{tabularray}
\usepackage{diagbox}
\usepackage{colortbl}
\usepackage{float}
\usepackage{placeins}
\usepackage[inline]{enumitem}
\usepackage{graphicx}

\renewcommand{\arraystretch}{1.1}

\def\BibTeX{{\rm B\kern-.05em{\sc i\kern-.025em b}\kern-.08em
    T\kern-.1667em\lower.7ex\hbox{E}\kern-.125emX}}

\begin{document}

\title {Towards Robust Robot 3D Perception in Urban Environments: The UT Campus Object Dataset}
\author{\IEEEauthorblockN{
Arthur Zhang, 
Chaitanya Eranki$^*$,
Christina Zhang$^*$,
Ji-Hwan Park$^*$,
Raymond Hong$^*$, 
Pranav Kalyani$^*$,\\
Lochana Kalyanaraman$^*$,
Arsh Gamare$^*$,
Arnav Bagad,
Maria Esteva,
Joydeep Biswas}%
\thanks{$^*$Equal Contribution.}
\thanks{A. Zhang, C. Eranki, C. Zhang, J. Park, R. Hong, P. Kalyani, L. Kalyanaraman, A. Gamare, A. Bagad, M. Esteva, and J. Biswas are with the University of Texas, Austin, TX 78705, USA. \tt\small{\{arthurz, chai77, yymzhang, raymond22, lochanak, arnavb, joydeepb\}@cs.utexas.edu} \tt\small{\{jihwanpark98, pranavkalyani, arsh.gamare\}@utexas.edu} \tt\small{maria@tacc.utexas.edu} }
}
\maketitle

\begin{abstract}
We introduce the UT Campus Object Dataset (\coda{}), a mobile robot egocentric perception dataset collected on the University of Texas Austin Campus. Our dataset contains 8.5 hours of multimodal sensor data: synchronized 3D point clouds and stereo RGB video from a 128-channel 3D LiDAR and two 1.25MP RGB cameras at 10 fps; RGB-D videos from an additional 0.5MP sensor at 7 fps, and a 9-DOF IMU sensor at 40 Hz. We provide 58 minutes of ground-truth annotations containing 1.3 million 3D bounding boxes with instance IDs for 53 semantic classes, 5000 frames of 3D semantic annotations for urban terrain, and pseudo-ground truth localization. We repeatedly traverse identical geographic locations for a wide range of indoor and outdoor areas, weather conditions, and times of the day. Using \coda{}, we empirically demonstrate that: 1) 3D object detection performance in urban settings is significantly higher when trained using \coda{} compared to existing datasets even when employing state-of-the-art domain adaptation approaches, 2) sensor-specific fine-tuning improves 3D object detection accuracy and 3) pretraining on \coda{} improves cross-dataset 3D object detection performance in urban settings compared to pretraining on AV datasets. Using our dataset and annotations, we release benchmarks for 3D object detection and 3D semantic segmentation using established metrics. In the future, the \coda{} benchmark will include additional tasks like unsupervised object discovery and re-identification. We publicly release \coda{} on the~\href{https://doi.org/10.18738/T8/BBOQMV}{Texas Data Repository}~\cite{coda2023tdr}, \href{https://github.com/ut-amrl/coda-models}{pre-trained models}, \href{https://github.com/ut-amrl/coda-devkit}{dataset development package}, and interactive dataset viewer\footnote{Interactive dataset viewer available on the \href{https://amrl.cs.utexas.edu/coda}{CODa website}}. We expect \coda{} to be a valuable dataset for research in egocentric 3D perception and planning for autonomous navigation in urban environments.

\end{abstract}


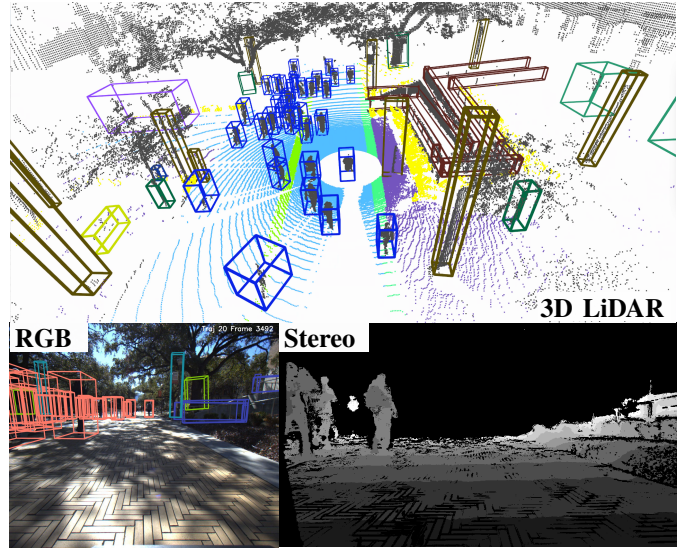
\begin{figure}[htbp]
\centering
\def\svgwidth{1.0\linewidth}
\begingroup%
  \makeatletter%
  \providecommand\color[2][]{%
    \errmessage{(Inkscape) Color is used for the text in Inkscape, but the package 'color.sty' is not loaded}%
    \renewcommand\color[2][]{}%
  }%
  \providecommand\transparent[1]{%
    \errmessage{(Inkscape) Transparency is used (non-zero) for the text in Inkscape, but the package 'transparent.sty' is not loaded}%
    \renewcommand\transparent[1]{}%
  }%
  \providecommand\rotatebox[2]{#2}%
  \newcommand*\fsize{\dimexpr\f@size pt\relax}%
  \newcommand*\lineheight[1]{\fontsize{\fsize}{#1\fsize}\selectfont}%
  \ifx\svgwidth\undefined%
    \setlength{\unitlength}{1585.86686262bp}%
    \ifx\svgscale\undefined%
      \relax%
    \else%
      \setlength{\unitlength}{\unitlength * \real{\svgscale}}%
    \fi%
  \else%
    \setlength{\unitlength}{\svgwidth}%
  \fi%
  \global\let\svgwidth\undefined%
  \global\let\svgscale\undefined%
  \makeatother%
  \begin{picture}(1,0.81784938)%
    \lineheight{1}%
    \setlength\tabcolsep{0pt}%
    \put(0,0){\includegraphics[width=\unitlength,page=1]{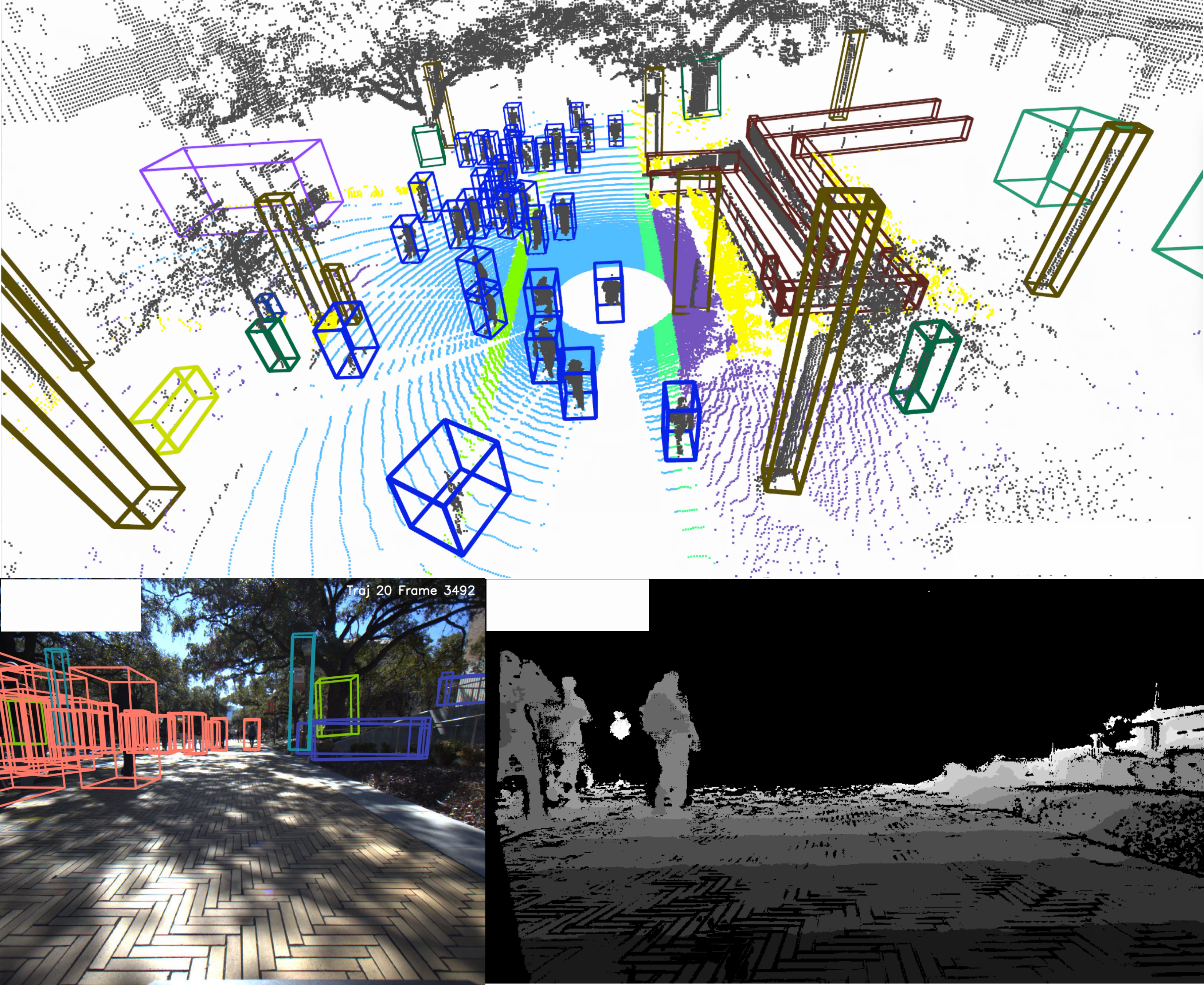}}%
    \put(0.00766798,0.30047644){\color[rgb]{0,0,0}\makebox(0,0)[lt]{\lineheight{1.25}\smash{\begin{tabular}[t]{l}\textbf{RGB}\end{tabular}}}}%
    \put(0.7966367,0.34555591){\color[rgb]{0,0,0}\makebox(0,0)[lt]{\lineheight{1.25}\smash{\begin{tabular}[t]{l}\textbf{3D LiDAR}\end{tabular}}}}%
    \put(0.41157487,0.30112535){\color[rgb]{0,0,0}\makebox(0,0)[lt]{\lineheight{1.25}\smash{\begin{tabular}[t]{l}\textbf{Stereo}\end{tabular}}}}%
  \end{picture}%
\endgroup%

\caption{Three of the five modalities available in \coda{}. \textbf{RGB} image with \textbf{3D object} labels (bottom left), \textbf{3D point cloud} (middle), \textbf{stereo depth} image (bottom right).
}
\figlabel{coda-cover}
\end{figure}

\section{Introduction}

Accurate and robust perception of objects and scenes is crucial for autonomous mobile robots performing tasks in urban environments. To this end, the computer vision and robotics communities have proposed datasets and benchmarks~\cite{MScoco, PascalVOC, ImageNet, KITTIVisionSuite, SCAND, JRDB} to serve as training data for the development and fair evaluation of modern data-driven approaches. However, perception models trained on existing datasets do not perform well in urban environments for the following reasons: 1)~they exhibit significant sensor and viewpoint differences from urban robots, 2)~they focus exclusively on RGB images, 3)~they lack sufficient object or terrain annotation diversity. These characteristics limit egocentric robot capabilities~\cite{panopticsegmentation, GANav, 3dreconstruction}, which are important for navigation and planning tasks.

Many egocentric 3D perception datasets are collected from urban robots or autonomous vehicles (AVs). Existing urban robotics datasets~\cite{SCAND, JRDB, NCLT} in human-centric environments possess similar sensors and viewpoints but lack semantic annotation diversity. In contrast, autonomous vehicle (AV) datasets~\cite{KITTIVisionSuite, Waymo, nuscenes} contain semantic annotations but are collected from cars on streets, roads, or highways. They operate higher fidelity sensor suites, encounter different geometric and semantic entities, and have different sensor viewpoints compared to urban robots. This causes perception models trained on AV datasets to perform poorly on robots in urban settings --- \secref{av_adaptation} presents quantitative analyses demonstrating this significant performance gap.

To address this gap, we contribute the \textbf{UT Campus Object Dataset (\coda{})}, a large-scale annotated multimodal dataset for training and benchmarking egocentric 3D perception for robots in urban environments. Our dataset is comprised of 23 sequences in indoor and outdoor settings on a university campus and contains repeated traversals from different viewpoints, weather conditions (\textbf{sunny, rainy, cloudy, low-light}), and scene densities.

The sensor data includes 
\begin{enumerate*}[label=\arabic*)]
 \item 3D point clouds from a 128-channel 3D LiDAR,
 \item RGB images from a stereo camera pair synchronized with the 3D LiDAR,
 \item RGB-D images from an active depth camera, 
 \item RGB-D images from a passive depth camera, and 
 \item 9 DoF inertial measurements. 
\end{enumerate*}
The dataset includes sensor intrinsic and extrinsic calibrations for each sequence and pseudo ground truth global poses. 

\coda{} contains \textbf{1.3 million} ground truth 3D bounding box annotations, instance IDs, and occlusion values for objects in the 3D point cloud. Furthermore, it includes \textbf{5000} frames of 3D terrain segmentation annotations for 3D point clouds. All annotations are provided by human annotators, and labeled at 10Hz for 3D bounding boxes, and 2-10Hz  for terrain semantic segmentation. Compared to similar 3D perception datasets, \coda{} has far more class diversity, containing \textbf{53} object classes and \textbf{23} urban terrain types. This includes classes that are useful to urban navigation, such as doors, railings, stairs, emergency phones, and signs. Using our annotations, we release benchmarks using established metrics~\cite{KITTIVisionSuite},~\cite{SemanticKitti} for 3D object detection and 3D semantic segmentation with plans for perception and planning tasks relevant to autonomous navigation. 

In the rest of the manuscript, we review existing datasets and relate \coda{} to them (\secref{related_work}), describe the sensor setup (\secref{setup}), data collection procedure (\secref{data_collection}), dataset contents (\secref{analysis}), and annotation details. We characterize the semantic composition of our dataset, proposed train/validation/test splits, and provide qualitative sensor data visualizations. Finally, in \secref{results} we empirically analyze how: using \coda{} improves object detection performance for robots in urban settings, different 3D LiDAR resolutions affect pre-trained object detector performance, and pre-training on \coda{} outperforms AV datasets in cross-dataset object detection on JRDB~\cite{JRDB}.

\begin{table*}[ht!]
\footnotesize
\centering

\begin{tabular}{c|c|cccc|cccc|ccc}
    \cellcolor{White}{\color{black}\textbf{Dataset}}&
    \cellcolor{White}{\color{black}\textbf{Pose}} &
    \cellcolor{White}{\color{black}\textbf{\#Cls}}&
    \cellcolor{White}{\color{black}\textbf{\#3D/2D Bbx}} &
    \cellcolor{White}{\color{black}\textbf{\#3D/2D Seg}} &
    \cellcolor{White}{\color{black}\shortstack{\textbf{Inst.}\\\textbf{Labels}}} &
    \cellcolor{White}{\color{black}\shortstack{\textbf{\#3D Ann.}\\ \textbf{Frames}}} &
    \cellcolor{White}{\color{black}\shortstack{\textbf{3D}\\ \textbf{Frames}}} &
    \cellcolor{White}{\color{black}\shortstack{\textbf{3D Pts/}\\ \textbf{Frame}}} &
    \cellcolor{White}{\color{black}\shortstack{\textbf{2D}\\ \textbf{Frames}}}&
    \cellcolor{White}{\color{black}\textbf{Time of Day}}&
    \cellcolor{White}{\color{black}\textbf{Night/Rain}}\\
    \hline
    KITTI\cite{KITTIVisionSuite} & G+R & 3 & 80K & 43K & Y & 15K & 15K & 120K & 13K & M, A & N/N\\
    nuScenes\cite{nuscenes} & G+R & 23 & 1.4M & 40K & Y & 40K & 400K & 34K & 1.4M & M, A, E & Y/Y\\
    Argoverse2\cite{Argoverse2} & G+R & 30 & 12M & 0 & Y & 150K & 6M & 107K & 300K & M, A, E & Y/Y\\
    Waymo Open\cite{Waymo} & G+R & 4 & 12M & 230K & Y & 192K & 192K & 177K & 1M & M, A, E & Y/Y\\
    ONCE\cite{once} & G & 5 & 417K & 0 & Y & 21K & 1M & 720K & 7M & M, A, E & Y/Y\\
    KITTI-360\cite{kitti360} & G+R & 14 & 68K & 156K & Y & 100K & 100K & ~200K & 150K & Not Given  & Not Given\\
    CityScapes3D\cite{Cityscapes3D} & G+R & 8 & Not Given & 20K & Y & 20K & 0 & N/A & 25K & M, A, E & Y/Y\\
    BDD100K\cite{BDD100K} & G+R & 10 & 1.8M$_{2D}$ & 10K$_{2D}$ & Y & 0 & 0 & N/A & 120M & M, A, E & Y/Y\\
    Oxford RobotCar\cite{RobotCarDatasetIJRR} & G+R & 0 & 0 & 0 & N & 0 & 0 & N/A & 20M & M, A, E & Y/Y\\
    ApolloCar3D\cite{apollocar3d} & G & 6 & 60K & 120K$_{2D}$ & Y & 0 & Not Given & N/A & 5.27K & M, A, E & Y/Y\\
    Lyft L5\cite{lyftl5} & G+R & 9 & 15K & 15K & Y & Not Given & Not Given & Not Given & 323K & M, A & N/N\\
    \hline 
    \coda{} (Ours) & \shortstack{S+G} & 53 & 1.1M & 6K & Y & 32K & 324K & 131K & 131K & M, A, E & Y/Y\\
\end{tabular}

\caption{Comparing dataset statistics between \coda{} (ours) and existing AV datasets. We use the following abbreviations: M - Morning, A - Afternoon, E - Evening, G - GPS, R - RTK, S - SLAM}
\tablabel{dataset_stats}
\end{table*}

\begin{table*}[ht!]
\footnotesize
\centering

\begin{tabular}{c|c|*{2}{c}cc|cccc|ccc}
    \cellcolor{White}{\color{black}\textbf{Dataset}}&
    \cellcolor{White}{\color{black}\textbf{Pose}} &
    \cellcolor{White}{\color{black}\textbf{\#Cls}} &
    \cellcolor{White}{\color{black}\textbf{\#3D Bbx}} &
    \cellcolor{White}{\color{black}\textbf{\#3D/2D Seg}} &
    \cellcolor{White}{\color{black}\shortstack{\textbf{Inst.}\\ \textbf{Labels}}} &
    \cellcolor{White}{\color{black}\shortstack{\textbf{\#3D Ann.}\\ \textbf{Frames}}} &
    \cellcolor{White}{\color{black}\shortstack{\textbf{3D}\\ \textbf{Frames}}} &
    \cellcolor{White}{\color{black}\shortstack{\textbf{3D Pts/}\\ \textbf{Frame}}} &
    \cellcolor{White}{\color{black}\shortstack{\textbf{2D}\\ \textbf{Frames}}} &
    \cellcolor{White}{\color{black}\textbf{In/Out}}&
    \cellcolor{White}{\color{black}\textbf{Time of Day}}&
    \cellcolor{White}{\color{black}\textbf{Night/Rain}}\\

    \hline

    MIT Stata\cite{mitstata} & S & 0 & 0 & 0 & Y & 0 & $\sim$5.1M & 1.4K & $\sim$5.1M & \makecell{I} & \makecell{N/A} & N/N\\
    TUM RGB-D\cite{sturm12iros} & MC & 0 & 0 & 0 & N & 0 & 47K & 0 & 47K & \makecell{I} & \makecell{N/A} & N/N\\
    Newer College\cite{newercollegedataset} & S & 0 & 0 & 0 & N & 0 & 23K & 131K & 23K & \makecell{I} & \makecell{M, A} & N/N\\
    JRDB\cite{JRDB} & None & 1 & \textbf{1.8M$^*$} & 0 & Y & 28K & 28K & 130K & 28K & \makecell{I+O} & \makecell{M, A} & N/N\\
    SCAND\cite{SCAND} & None & 0 & 0 & 0 & N & 0 & 313K & 65K & 626K & \makecell{I+O} & \makecell{M, A} & N/N\\
    RUGD\cite{RUGD2019IROS} & None & 24 & 0 & 7.4K$_{2D}$ & N & 0 & 0 & 0 & 37K & \makecell{O} & \makecell{M, A} & N/N\\
    Rellis-3D\cite{Rellis3D} & S+G & 20 & 0 & 13K & N & 13K & 13K & \textbf{1.33M} & 6K & \makecell{O} & \makecell{M, A} & N/N\\
    NCLT\cite{NCLT} & S+G+R & 0 & 0 & 0 & N & 0 & 1.2M & 69.5K & 628K & \makecell{I+O} & \makecell{M, A, E} & Y/N\\
    ALITA\cite{yin2022alita} & S & 0 & 0 & 0 & N & 0 & \textbf{7.2M} & 65K & \textbf{7.2M} & \makecell{O} & \makecell{M, A} & N/N\\
    FusionPortable\cite{fusionportable} & S+MC & 0 & 0 & 0 & N & 0 & 1.4M & 131K & 2.9M & \makecell{I+O} & \makecell{M, A} & N/N\\
    OpenLORIS\cite{shi2019openlorisscene} & MC & \textbf{40} & 0 & 0 & N & 0 & 497K & N/A & 497K & \makecell{I+O} & \makecell{M, A} & N/N\\
    Pascal VOC3D+\cite{xiang_wacv14} & None & 12 & 36K & 0 & Y & 30K & 0 & N/A & 22K & \makecell{I+O} & \makecell{M, A, E} & N/N\\
    NYU Depthv2\cite{nyudepthv2} & None & 26 & 0 & 1.45K  & Y & 1.45K & 407K & N/A & 407K & \makecell{I} & \makecell{M, A, E} & N/N\\
    \hline 
    \textbf{\coda{} (Ours)} & \textbf{S+G} & \textcolor{blue}{\textbf{53}} & \textcolor{blue}{\textbf{1.3M}} & \textbf{6K} & \textbf{Y} & \textcolor{blue}{\textbf{32K}} & \textbf{324K} & \textbf{131K} & \textbf{324K} & \textcolor{blue}{\textbf{I+O}} & \textcolor{blue}{\textbf{M, A, E}} & \textcolor{blue}{\textbf{Y/Y}}\\
\end{tabular}

\caption{Comparison between \coda{} (ours) and similar campus scale robot datasets. The most significant entry for each column in \coda{} is highlighted in blue. \coda{} provides the largest number of object classes, 3D bounding box annotations, and annotated 3D frames under the widest range of environmental and weather conditions.
Pose annotations: G - GPS, R - GPS-RTK, S - SLAM, MC - Motion Capture.
Indoor/Outdoor: I - Indoor, O - Outdoor. 
Time of Day: M - Morning, A - Afternoon, E - Evening.
* JRDB\cite{JRDB} only provides annotations for pedestrians.
}
\tablabel{campus_datasets}
\end{table*}


\section{Related Work}
\seclabel{related_work}
In this section, we review existing egocentric 3D LiDAR datasets for urban and AV domains. We limit the discussion to real-world datasets, as there still exists a significant domain gap between simulation and real-world~\cite{hoffman2016fcns,DBLP:journals/corr/abs-1802-10349}. 

\subsection{Urban Datasets}

Urban datasets are collected in human-centric environments, such as college campuses, city streets, and shopping malls. Similar to our work, these datasets are used to benchmark robot performance in human-centric environments, often emphasizing long-term SLAM, object detection, and semantic segmentation. While there exist computer vision benchmarks for 3D object detection~\cite{PascalVOC} and semantic segmentation~\cite{nyudepthv2}, we focus on datasets collected from mobile robots due to differences in perspective shift and sensor suite.

Long-term SLAM datasets like MIT Stata~\cite{mitstata}, NCLT~\cite{NCLT}, FusionPortable~\cite{fusionportable}, and OpenLORIS~\cite{shi2019openlorisscene} contain globally consistent ground truth poses and multimodal sensor data. They are repeatedly collected over multiple times of day to fairly evaluate long-term SLAM methods that rely on geometric, visual, or proprioceptive sensor information. SCAND~\cite{SCAND} is another large-scale dataset with multimodal sensor data collected over multiple times of day in a campus environment. Instead of ground truth poses, it contains socially compliant navigation demonstrations and operator commands to support social navigation research. Similarly, \coda{} contains multimodal sensor data with repeated trials over multiple times of day, but distinguishes itself by providing object and terrain annotations to support methods that rely on semantic information.

Besides \coda{}, there does not exist an urban robot dataset that contains 3D object and terrain annotations. RUGD~\cite{RUGD2019IROS} and Rellis-3D~\cite{Rellis3D} are robot datasets with 2D and 3D semantic segmentation annotations respectively, but are collected on off-road terrains. These environments contain distinct semantic entities from those found in urban environments. The closest work to ours is JRDB~\cite{JRDB}, a mobile robot dataset with 1) 1.8 million 3D bounding box annotations 2) indoor and outdoor sequences 3) egocentric sensor data. However, JRDB~\cite{JRDB} is intended for pedestrian understanding research as it only contains pedestrian semantic annotations. In contrast, \coda{} contains object and terrain level annotations for a wide range of semantic classes to support general-purpose egocentric perception and navigation in urban environments.

\subsection{AV Perception Datasets}

Unlike urban robot datasets, AV datasets are collected from car-mounted, high-fidelity sensor suites and operate exclusively on roads, parking lots, and highways. Despite these differences, their large size and scene diversity may be leveraged to train 3D perception algorithms for urban settings.

Among AV datasets, the Oxford RobotCar dataset~\cite{RobotCarDatasetIJRR} contains the most repeated traversals over different weather, object density, and lighting conditions. It provides ground truth poses for evaluating long-term SLAM methods that only rely on visual and geometric information. For 2D multitask learning problems, Berkeley DeepDrive~\cite{BDD100K} provides semantic annotations at both the object and pixel level for a wide range of semantic classes and weather conditions.

Lyft L5~\cite{lyftl5}, CityScapes3D~\cite{Cityscapes3D}, and KITTI-360~\cite{Liao2022PAMI} contain labeled 2D images or 3D bounding boxes with more non-overlapping semantic classes than other AV datasets. They contain vehicle-centric semantic classes to support multi-class object detection research in AV domains. Conversely, large-scale datasets like Waymo Open~\cite{Waymo} and nuScenes~\cite{nuscenes} have fewer unique semantic classes but have more 3D semantic annotations per class and greater scene diversity. These characteristics establish them as de facto benchmarks for 3D object detection and semantic segmentation tasks, while also being valuable for pre-training 3D object detectors to recognize similar objects across domains. 

Other works like Argoverse2~\cite{Argoverse2} and ONCE~\cite{once} support self-supervised point cloud learning by providing more unannotated 3D point clouds than any other AV dataset. Additionally, both contain 3D object labels, but Argoverse2's~\cite{Argoverse2} labels are limited to five meters within the drivable area and ONCE~\cite{once} is limited to five object classes. For robots operating in urban environments, it is important to identify a diverse set of objects in non-drivable areas, reinforcing the need for a dataset like \coda{}.

\begin{figure}[htbp]
\centering
\includegraphics[width=0.48\textwidth, trim=25 0 150 50, clip]{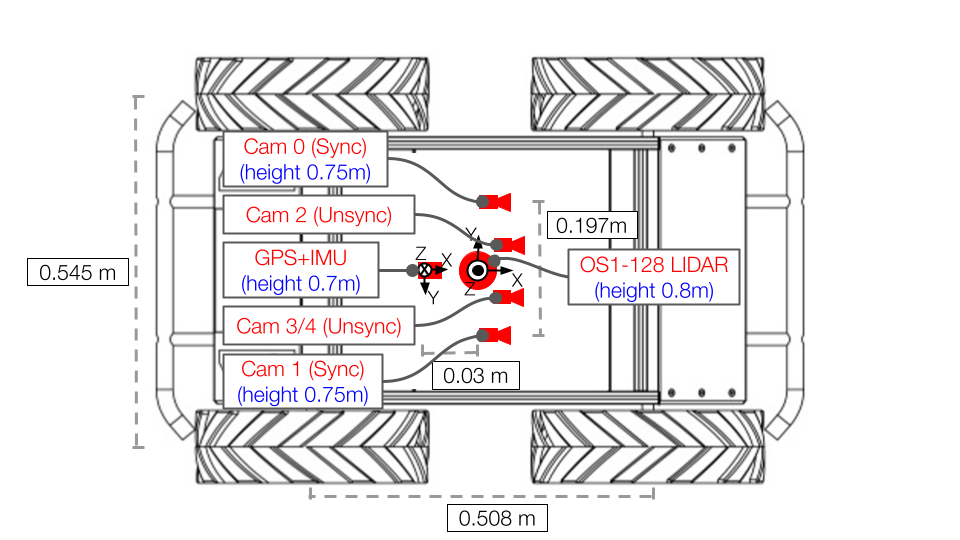}
\caption{\textbf{Sensor setup} including mounting positions. All heights are relative to the ground plane.}
\figlabel{husky-sensors}
\end{figure}

\section{Sensor Setup}
\seclabel{setup}
\coda{} was collected using a Clearpath Husky robot~\cite{clearpath-husky} equipped with a custom sensor suite with the following sensors, illustrated in \figref{husky-sensors}:
\begin{itemize}
    \item 1 $\times$ Ouster OS1-128 3D LIDAR, 128 beams - $0.35\degree$ vertical angular resolution, 2048 beams - $0.17\degree$ horizontal angular resolution, up to 2.6 million points/second, field of view: 360$\degree$ horizontal, 45$\degree$ vertical, range: 128 m. Point clouds captured in 128x1024 channels @ 10 Hz.
    \item 2 $\times$ Teledyne FLIR Blackfly S RGB cameras (BFS-U3-51S5C-C) up to 5 Megapixels, 75 Hz, global shutter. Paired with KOWA F2.8/5mm lenses. Field of view (H x W): 70$\degree$x79$\degree$. Images captured in 1.25 Megapixels @ 10 Hz, hardware synchronized with 3D LIDAR.
    \item 1 $\times$ Microsoft Azure Kinect active RGBD camera up to 12 and 1 MP (RGB and Depth) @ 15 Hz, rolling shutter. 7 microphone circular array. RGB and Depth Images captured in 2.0 MP @ 5Hz
    \item 1 $\times$ Stereolabs ZED 2i passive stereo camera up to 4 Megapixels @ 15 Hz, rolling shutter. Images captured in 0.5MP @ 5Hz
    \item 1 $\times$ Vectornav VN-310 Dual GNSS/INS, up to 800 Hz IMU Data. Inertial and GPS data captured @ 40Hz
\end{itemize}
An onboard computer with an Intel I7-8700 3.2GHz CPU and 32 GB RAM is securely mounted inside the robot and records all sensor streams to a high-speed Intel 760P SSDPEKKW512G8 512GB SSD. A GPU-equipped laptop is mounted on the robot to process the Azure Kinect and ZED 2i camera data before transmitting both RGBD streams to the computer via 10 Gig Ethernet. The coordinate system definitions are described in the \coda{} documentation~\cite{coda2023tdr}.

The Ouster LIDAR and FLIR cameras are synchronized by hardware using the Ouster LIDAR 10Hz sync pulse to trigger the FLIR cameras. This ensures that the start of the LIDAR scan is synchronized with the start of the exposure of the FLIR cameras. All other sensors have timestamps, but their capture times are not synchronized.

We calibrate the stereo RGB cameras with a checkerboard calibration pattern~\cite{zhang2000calibration} using multiple images of the checkerboard at different positions. We obtain the LiDAR camera extrinsics using checkerboard images and an approach~\cite{tsai2021optimising} that optimizes the sensor pose with respect to the checkerboard target and the entire scene. To obtain the LiDAR-IMU extrinsic, we use a target-free extrinsic calibration algorithm~\cite{mishralidarimucalibration2021} that exploits vehicle motion. A calibration half-cube is used to ensure that the LiDAR depth camera extrinsic is accurate. Every sequence in \coda{} includes a pre or post-run calibration log file containing the raw sensor data and calibration targets in the field of view. \figref{calibration} shows a sample frame from the calibration log file. 

\begin{figure}[h]
    \centering
    \includegraphics[width=\linewidth]{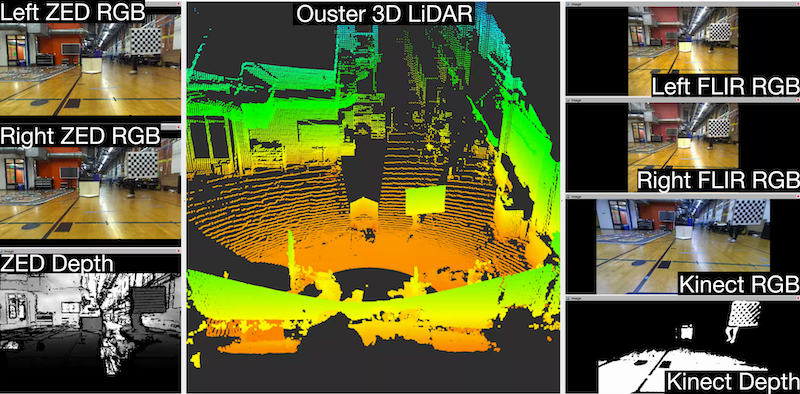}
    \caption{Sample frame from calibration file. Calibration half-cube and checkerboard are simultaneously visible in all RGB, depth, and 3D LIDAR frames.}
    \figlabel{calibration}
\end{figure}

\section{Data Collection Procedure}
\seclabel{data_collection}
In this section, we describe the sensor calibration procedure and data collection routes for \coda{}.

\subsection{Per Sequence Calibration Procedure}
\seclabel{perseqcalibrationprocedures}

\begin{table}[h]
    \centering
    \begin{tabular}{l|l|l|l|r|r|r}
    \textbf{Route} & \textbf{Setting} & \textbf{Locations} & \textbf{Traversals} & \textbf{Dist. (m)} & \textbf{Dur. (hr)} \\
    \hline
    Gates-Dell & Both & GDC, SWY & 7 & $5139$ & $1.95$ \\
    Guad24 & Out & Guad, SWY & 6 & $8799$ & $3.07$ \\
    WCPowers & Both & WCP, SWY & 7 & $8005$ & $2.79$ \\
    Union & In & UNB, Guad & 3 & $2450$ & $0.93$ \\
    \end{tabular}
    \caption{Summary of the four routes in \coda{}. We traverse each route multiple times to capture diverse viewpoints, weather, and lighting conditions. Each route passes through a set of geographic locations defined in \figref{campusmap}. The setting column describes whether the route is indoors, outdoors, or both. }
    \tablabel{routes}
\end{table}

\begin{figure}[ht]
    \centering
    \includegraphics[width=\linewidth]{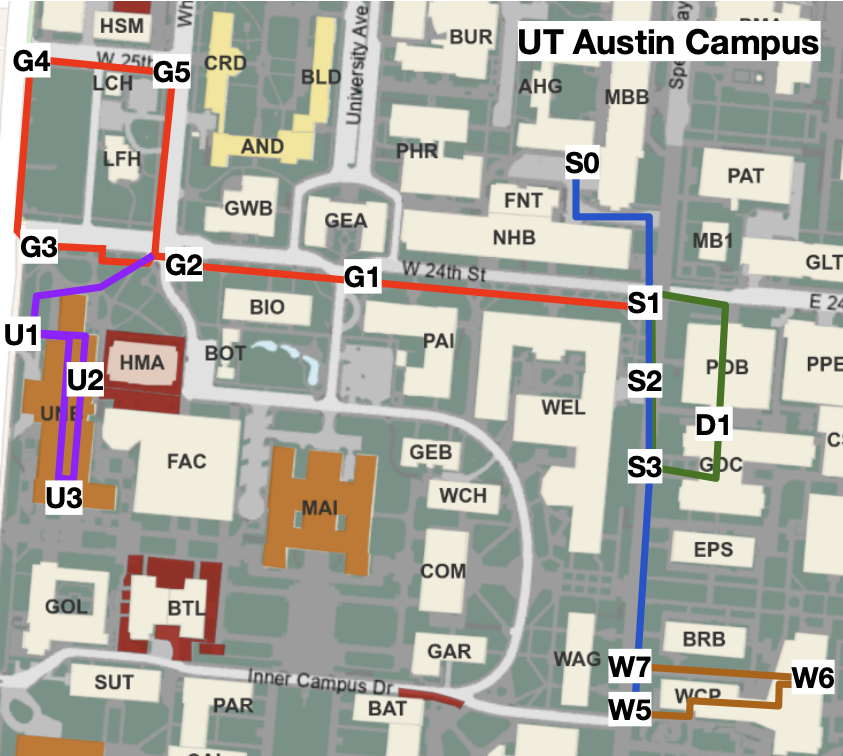}
    \caption{Spatial map of geographic locations contained in \coda. Operators pause the robot at each waypoint denoted on the map to correct global pseudo-ground truth pose estimates. We refer to blue, green, brown, red, and purple locations as SWY, GDC, WCP, Guad, and UNB in \figref{locationweathercount} }
    \figlabel{campusmap}
\end{figure}

Robot operators calibrated the 3D LiDAR and RGB cameras using the methods in \secref{setup} for each sequence and saved the raw sensor data into a calibration log. Because the 3D LiDAR and IMU are fixed with respect to each other, we performed the LiDAR-IMU calibration once and used this calibration for all sequences. We moved the checkerboard at three different heights across the image to obtain accurate stereo camera calibrations. \figref{calibration} shows the calibration process and sensor modalities in the calibration log. 

\subsection{Operator Roles and Data Privacy}

After calibration, pairs of operators drove the robot along one of four
predetermined routes on UT campus. The primary operator drove the robot along 
predefined routes, including stopping at waypoints defined in \figref{campusmap}, which are used to ensure
global pose consistency between sequences. The second operator addressed
questions from the crowd about \coda{} and handed out research information
sheets containing a data privacy disclaimer and contact information. This
operator logged all individuals' requests to opt-out from participating in
\coda{}. We increase transparency by mounting a sign on the robot to indicate
when it is recording data. While no individuals opt-ed out from our experiments,
we protect the privacy of those who do with our user data removal procedure. We
describe the user data removal procedure and release the research information
sheet on TDR~\cite{coda2023tdr}. In the next section, we explain the routes in
detail. 

\subsection{Data Collection Routes}
\seclabel{routesprocedures}

The four navigation routes along UT campus are: \textsc{Gates-Dell}, \textsc{WCPowers}, \textsc{Guad24}, and \textsc{Union}. \figref{satelliteglobalmap} shows the reconstructed map for the first three routes in red, blue, and green. We summarize the characteristics for each route in \tabref{routes}, including the total distance traversed, total duration, number of traversals, and geographic locations visited for each route. These geographic locations are shown in \figref{campusmap} as \textsc{SWY}, \textsc{GDC}, \textsc{WCP}, \textsc{Guad}, and \textsc{UNB}. We choose each location for the following attributes: \textsc{SWY} has a large open area shared by vehicles and pedestrians, \textsc{GDC} has large open areas with classrooms, \textsc{WCP} has scenes from a cafeteria, \textsc{Guad} has scenes from sidewalks and vehicle-only roads, and \textsc{UNB} has scenes from a library and study area.

Each location is observed multiple times from various viewpoints, weather, and lighting conditions. We quantify the observation diversity in \figref{locationweathercount} by counting the number of observed frames in \coda{} for each location under four weather/lighting conditions (\textsc{Cloudy}, \textsc{Dark}, \textsc{Sunny}, \textsc{Rainy}) during three times of day (\textsc{Morning, Afternoon, Evening}). While we are unable to deploy the robot when it is actively raining, we collect data immediately after rainfall and label frames that satisfy these conditions as rainy. Across all sequences, \coda{} contains 3 rainy, 7 cloudy, 4 dark, and 9 sunny sequences. \figref{coda_gt_visualizations_part1} qualitatively showcases the data diversity in \coda{} using sampled images from each sequence.

\begin{figure}[ht]
    \centering
    \includegraphics[width=\linewidth]{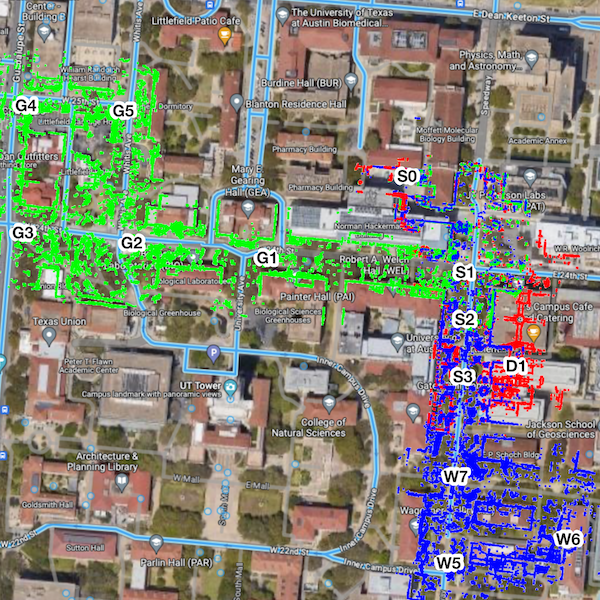}
    \caption{Satellite image of UT campus with the transformed point clouds and waypoints overlaid. The blue, green, and blue points correspond to the WCP, GDC, and Guad routes respectively. Operators pause the robot at each waypoint to establish global correspondences for pseudo-ground truth pose estimates. Most sequences exhibit poor GPS reception, thus requiring poses to be estimated from LiDAR, inertial, and waypoint data. }
    \figlabel{satelliteglobalmap}
\end{figure}
\begin{figure}[ht]
    \centering
    \includegraphics[width=\linewidth]{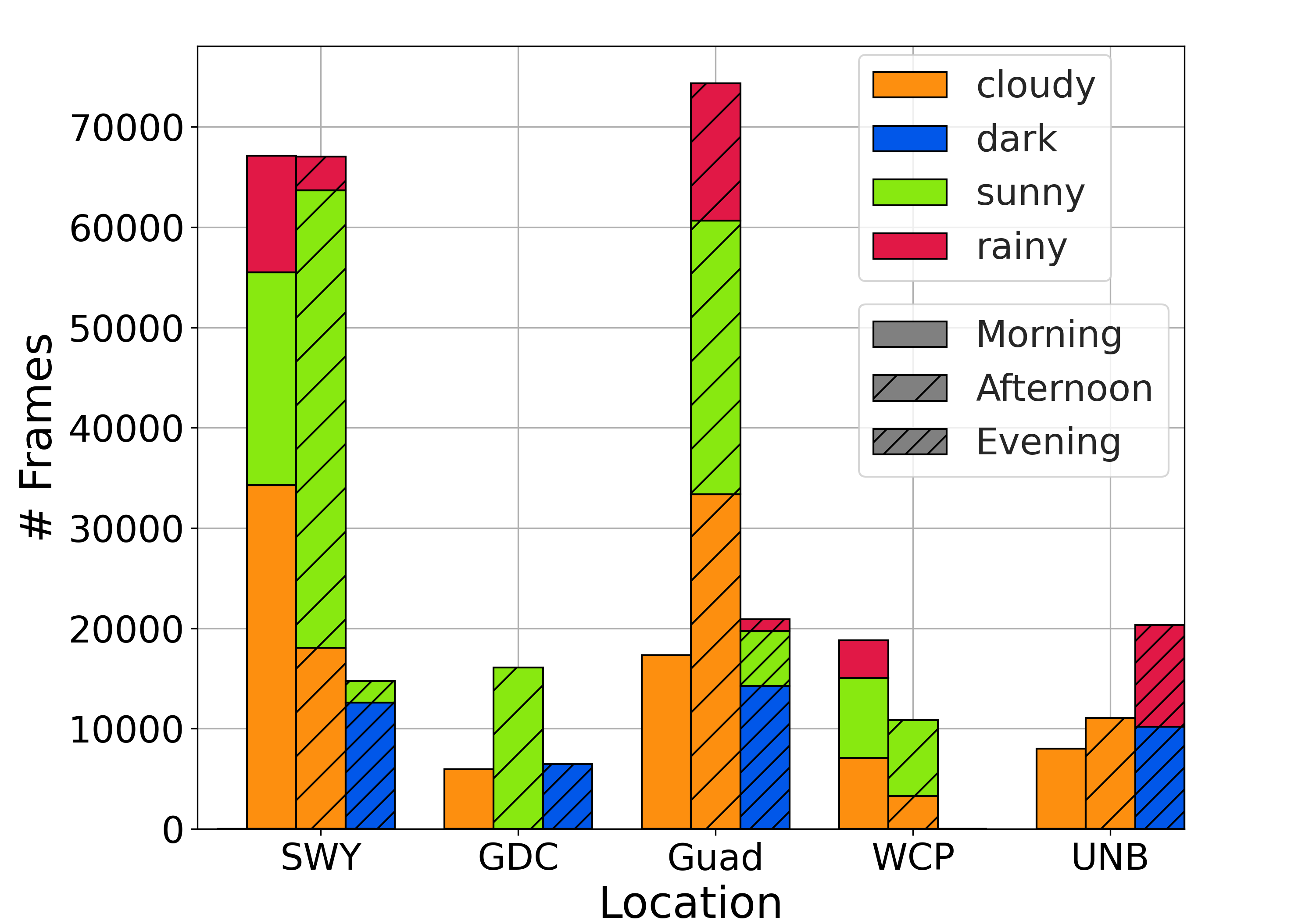}
    \caption{Number of frames in \coda{} by geographic location and weather condition. Locations with temporally diverse observations contain frames during multiple times of the day. The coverage areas for locations SWY, GDC, WCP, Guad, and UNB are marked by blue, green, brown, red, and purple lines respectively in \figref{campusmap}.}
    \figlabel{locationweathercount}
\end{figure}

\section{Annotation and Labels}
\seclabel{annotation}
We utilized Deepen AI \footnote{Company Website (Deepen): https://www.deepen.ai/}, a 3rd party annotation company, to annotate point clouds from our 3D LiDAR with 3D bounding box and semantic segmentation annotations. We instructed Deepen annotators using our annotation guide, which we provide in the data report~\cite{coda2023tdr}. The annotation guide contains visual examples for each object and terrain class described in \figref{fullobjectlist} and \figref{fullsemseglist}, quantitative occlusion level definitions, and operating procedures to determine object instance IDs. Following these instructions, Deepen annotators manually labeled 58 minutes of frames, followed by manual quality assurance checks to ensure that at least 95\% of the bounding boxes and 90\% of the terrain segmentation annotations were valid on the 3D point clouds. Our internal team then inspected each frame for additional issues. We now describe each annotation type in \coda{} in detail.

\subsection{3D Bounding Boxes}

Each 3D bounding box has 9 degrees of freedom, instance ID, object class, and occlusion level attributes. We maintain the same instance ID for each object as long as it is observable from the LiDAR or camera sensor or if it does not leave view for longer than 3 seconds. There are six occlusion types, ranging from \textsc{None}, \textsc{Light}, \textsc{Medium}, \textsc{Heavy}, \textsc{Full}, and \textsc{Unknown} occlusion. The first five occlusion types are used if the object is observable by the cameras or can be identified fully in the 3D point cloud. Objects that never enter the camera view or are geometrically ambiguous are given the unknown occlusion status. This label definition makes \coda{} useful for evaluating the 3D object tracking task under occlusion. \figref{fullobjectlist} defines the object ontology for \coda{}. Because the full list of object classes is large, we refer the reader to the data report~\cite{coda2023tdr} for visual examples of each class.

\subsection{3D Semantic Segmentation}

We annotate each point on the surrounding terrain with a semantic class label. We differentiate terrain classes by their visual appearance and geometric shape. For instance, red and yellow bricks are similar geometrically but are treated as different terrains because they are visually distinct. This makes 3D semantic segmentation challenging with just a single 3D LiDAR and encourages multi-modal methods that fuse 2D images and 3D LiDAR to infer terrain-level semantic labels. We label ambiguous points as unknown and points not associated with terrain as unlabeled. The full terrain ontology and examples for each class can be found in \figref{fullsemseglist} and \figref{semantic2dexamples} respectively.

\subsection{Pseudo Ground Truth Poses}

Due to the unreliability of GPS in urban environments, we use Lego-LOAM~\cite{legoloam2018} to obtain initial robot poses and HitL-SLAM~\cite{hitlslam} to refine these pose estimates globally between runs using manual annotations with known map correspondences. In \figref{campusmap}, we qualitatively assess our method's accuracy by visualizing the global pose estimate on a satellite image of UT campus and 3D map reconstruction. 

\section{Analysis of \coda{} Annotations and Statistics}
\seclabel{analysis}
In this section, we analyze the distribution of data in \coda{} by geographic location, weather, and lighting conditions. 

\begin{figure}
    \centering
    \includegraphics[width=\linewidth]{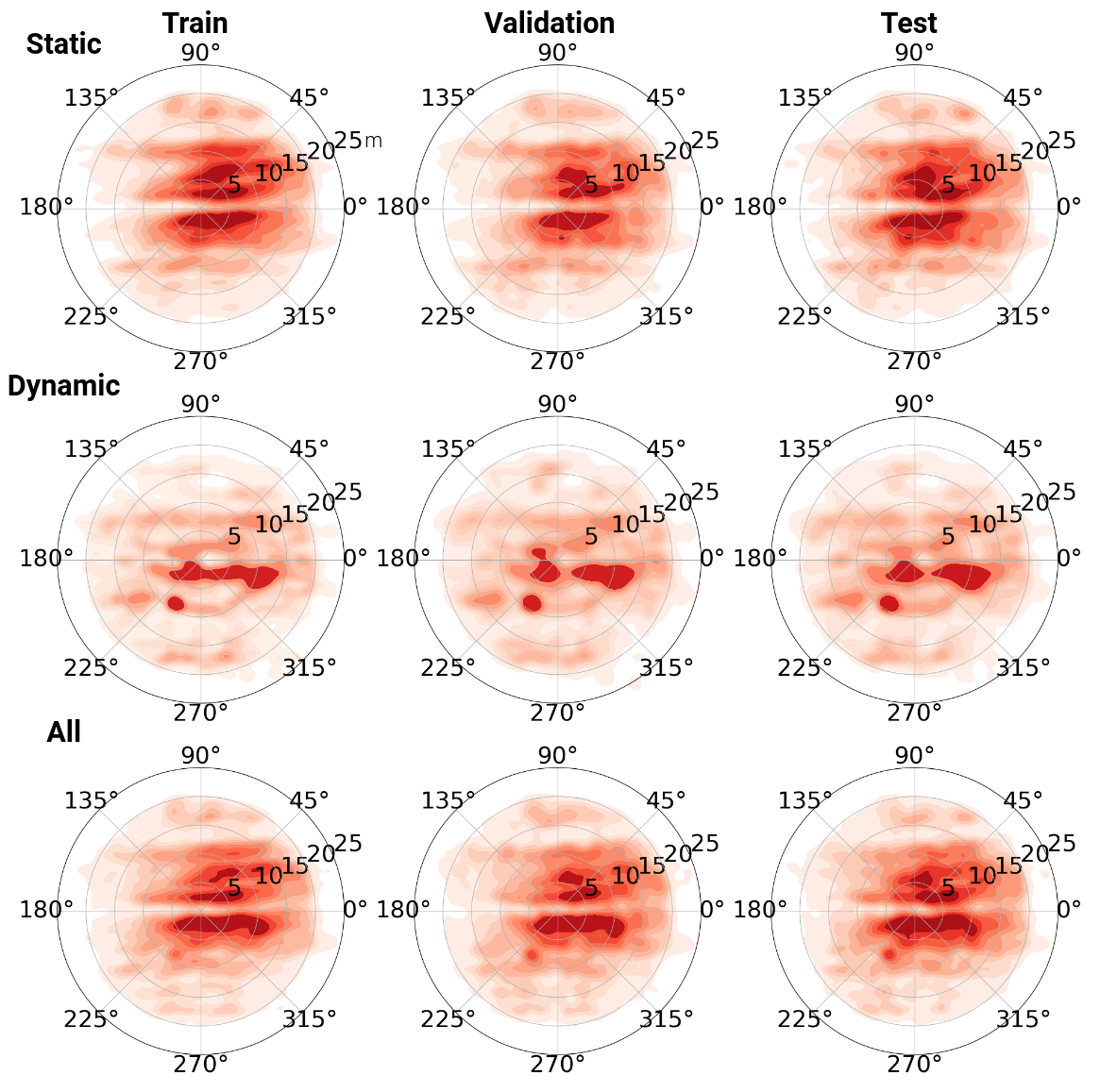}
    \caption{Spatial distribution of static (top), dynamic (middle), and all (bottom) objects around the robot for the train (left), validation(center), and test (right) splits. Angles (in degrees) are with respect to the forward heading of the robot, range values in meters.}
    \figlabel{objectheatmap}
\end{figure}

\begin{figure*}
    \centering
    \includegraphics[width=\linewidth]{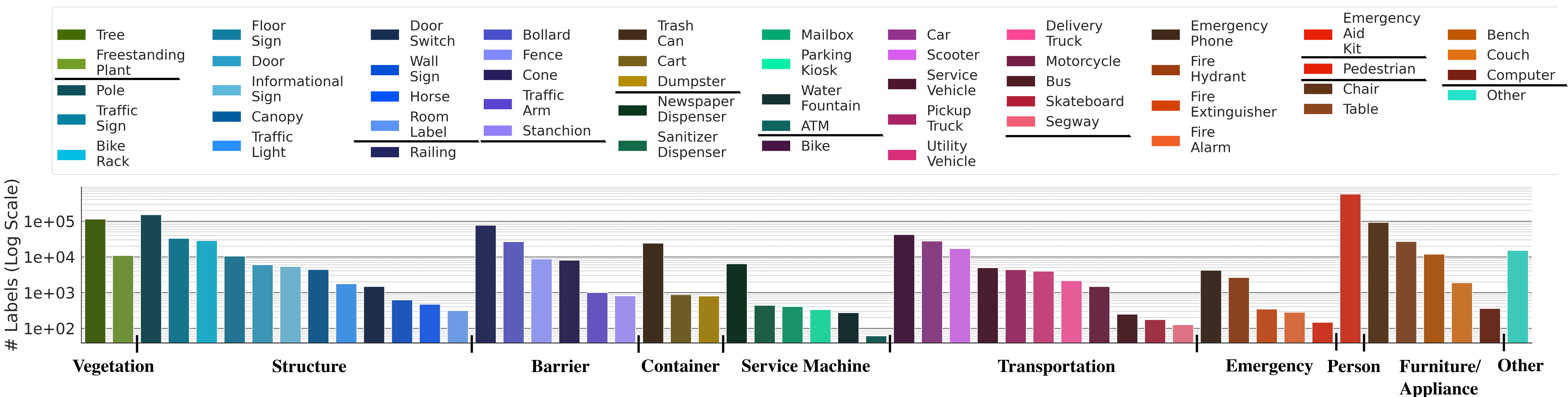}
    \caption{Number of object labels per class organized by topological category. Objects in each topological category are sorted in order of most to least common.}
    \figlabel{distributionobjects}
\end{figure*}

\begin{figure*}
    \centering
    \includegraphics[width=\linewidth]{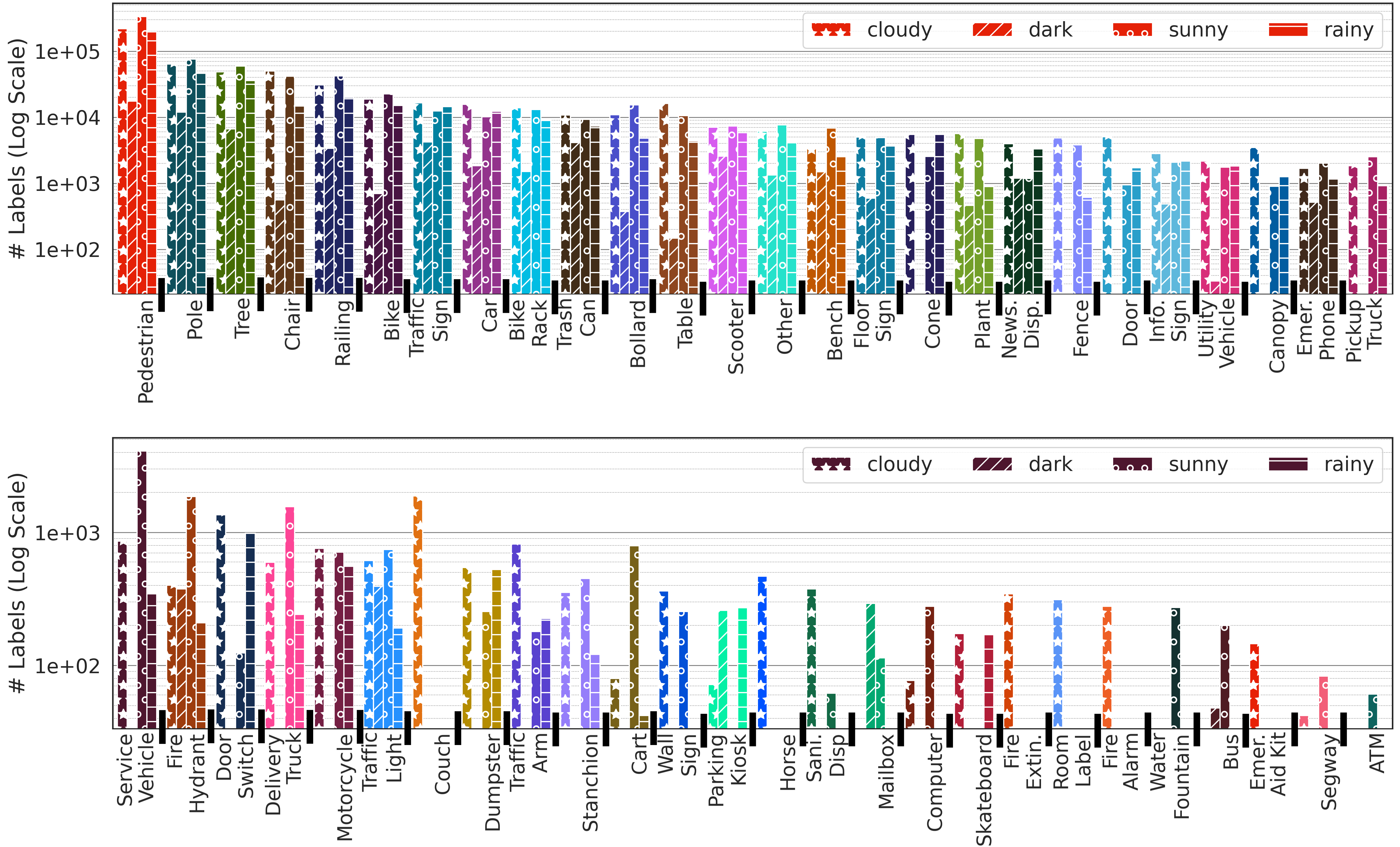}
    \caption{Histogram of the number of annotations per object class under four weather conditions (sunny, rainy, cloudy, dark). Object classes are organized by most to least frequent from left to right. Bars with stars are cloudy, diagonal lines are dark, circles are sunny, and horizontal lines are rainy.
    }
    \figlabel{objecthistogram}
\end{figure*}

\begin{figure}
    \centering
    \includegraphics[width=\linewidth]{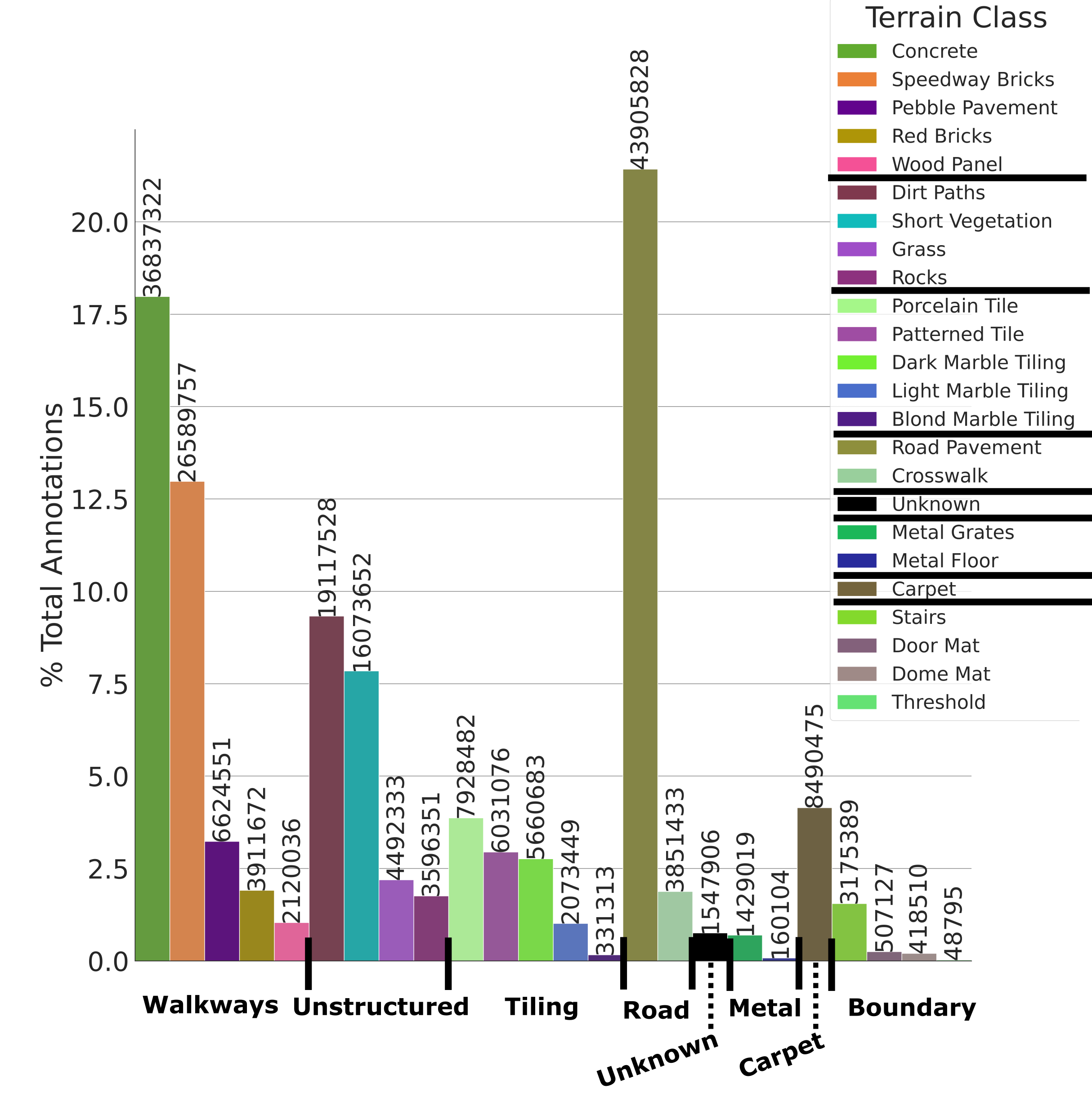}
    \caption{Histogram of 3D semantic segmentation annotations labels for outdoor, indoor, and both environments in \coda{}. Vertical numbers above each bar indicate the total number of points annotated for that semantic class. The semantic classes in the legend map to the bars from left to right.}
    \figlabel{semantichistogram}
\end{figure}

\figref{locationweathercount} shows that all geographic locations in \coda{} (besides WCP) contain data in the morning, afternoon, and evening. All routes with outdoor observations contain at least one sequence captured under rainy conditions. While the full dataset is biased toward sunny and cloudy weather, \figref{objecthistogram} shows that the annotated dataset contains 20 object classes that have at least 100 labels under all conditions. With this number going up to 36 classes if we only consider 3 of the 4 conditions. Aside from ATM, most classes contain 100 to 1000 labels each, with \figref{objecthistogram} showing the top five classes: pedestrian, tree, pole, railing, and chair. This class and weather imbalance is common in real-world datasets~\cite{nuscenes, once} and is a challenging aspect that perception algorithms deployed in urban environments need to be resilient to. 

\figref{semantichistogram} shows the proportion of each terrain class among the annotated points in \coda{}, organized by the parent class defined in the terrain ontology in \figref{fullsemseglist}. Among the 23 terrain classes, 21 have more than 200,000 annotated points each, with outdoor classes dominating the majority of the annotations. The two classes that do not satisfy this are dome mat and metal floor. This is because these terrains are small in size and uncommon in environments where they are found. This class imbalance is present in other real-world semantic segmentation datasets~\cite{SemanticKitti},~\cite{Cityscapes3D} as well.

We propose train, validation, and test splits for our 3D object detection and 3D terrain segmentation benchmarks, with each split containing 70\%, 15\%, and 15\% of each annotated sequence respectively. We visualize the spatial distribution of objects around the robot in \figref{objectheatmap} for static, dynamic, and all objects for each proposed split in a Kernel Density Estimate (KDE) plot. This demonstrates that both the density and relative position of objects around the robot are similar between our proposed splits.

\section{Experiments and Analysis}
\seclabel{results}
We leveraged the unique characteristics of \coda{} to conduct experiments that answer the following questions:

\begin{itemize}
    \item Question 1: How well do 3D object detectors trained on large-scale AV datasets perform on \coda{}?
    \item Question 2: How well does unsupervised domain adaptation from AV datasets perform on \coda{}? 
    \item Question 3: Can we improve object detection performance for low-resolution single 3D LiDAR setups on robots by fine-tuning on downsampled LiDAR point clouds?
    \item Question 4: Does pre-training on \coda{} improve cross-dataset object detection on existing urban robotics datasets?
\end{itemize}

\subsection{Experimental Setup --- Selecting a 3D Object Detection Algorithm}

We choose a 3D object detector by evaluating the performance of three 3D object detectors: PointPillars~\cite{lang2019pointpillars}, CenterPoint~\cite{centerpoint}. PVRCNN~\cite{pvrcnn} on KITTI, Waymo, nuScenes, and \coda{}. These datasets are among the most widely used in 3D object detection benchmarks and for cross-dataset domain adaptation analysis~\cite{2020objdetgeneralize, once}. We evaluate the preceding models because they are LiDAR-only approaches, easy to reproduce, and achieve state-of-the-art detection performance on AV datasets. Both Centerpoint and PointPillars are top-performing open-source methods on Waymo and nuScenes leaderboards, and the OpenPCDet~\cite{openpcdet2020} implementation of PVRCNN unofficially outperforms the former models on Waymo. We use the OpenPCDet implementation of each model because it provides the model configuration files, making results more reproducible.

We use the default model configurations provided in OpenPCDet and train each model for 30 epochs or until the performance saturates. For models that OpenPCDet does not provide configurations for, we benchmark various model architectures in \tabref{base_model_selection} and select the most favorable one. 

All experiments involving \coda{} in Tables~\ref{tab:base_model_selection},~\ref{tab:da_analysis}, and~\ref{tab:lidar_resolution_analysis} are conducted using the medium train, validation, and test split for computational reasons. We use the full \coda{} split for \tabref{jrdb_zeroshot_analysis} experiments to better match the scene diversity in AV datasets. For all metrics, we use the 3D object detection and bird's eye view evaluation metric proposed in the KITTI Vision Benchmark Suite~\cite{KITTIVisionSuite} with an IOU of 0.7, 0.5, and 0.5 for the car, pedestrian, and cyclist classes respectively. This class list is consistent across \coda{} and AV datasets. For completeness, we report model performance on the full list of object classes for multiple LiDAR resolutions in the appendix.

\begin{table}[ht]
\footnotesize
\centering

\renewcommand{\arraystretch}{1.25} 


\begin{tabular}{c|cc|cc|cc|cc}
    \cellcolor{White}{\color{black}\diagbox{\textbf{Mod.}}{\textbf{Data.}}}&
    \multicolumn{2}{c|}{\cellcolor{White}{\color{black}\textbf{nuScenes}}} &
    \multicolumn{2}{c|}{\cellcolor{White}{\color{black}\textbf{Waymo}}} &
    \multicolumn{2}{c|}{\cellcolor{White}{\color{black}\textbf{KITTI}}} &
    \multicolumn{2}{c}{\cellcolor{White}{\color{black}\textbf{\coda{}}}}\\
    \hline
    PointPillars~\cite{lang2019pointpillars} & 28.42 & 17.94 & 55.11 & 47.55 & \bestbev{70.27} & 63.32 & 49.78 & 48.86\\
    CenterPoint~\cite{centerpoint} & \bestbev{36.91} & 23.86  & 62.66 & 54.86 & 69.34 & 63.87 & 82.08 & 76.92\\
    PVRCNN~\cite{pvrcnn} & 33.85 &  \besttred{25.41} & \bestbev{62.73} & \besttred{56.40} & 70.22 & \besttred{65.28} & \bestbev{92.08} & \besttred{91.11}\\
\end{tabular}

\caption{Evaluation of several 3D object detectors on AV Datasets and \coda{}, We report mean average precision for the car, pedestrian, and cyclist categories in bird's eye view (AP$_{BEV}$) and 3D (AP$_{3D}$) with IOU 0.7, 0.5, and 0.5 respectively. We average the results at the easy, medium, and hard difficulties (following the KITTI Vision Benchmark). The blue and red indicate the highest-performing training method for BEV and 3D detection for each dataset. Mod. - Model Data. - Dataset} 

\tablabel{base_model_selection}
\end{table}

We observe in \tabref{base_model_selection} that PVRCNN generally performs the best for 3D bounding box detection on large-scale AV datasets and \coda{}. As such, we select this model architecture to use in all of our later experiments. For a full summary of all models evaluated for this experiment, please refer to \tabref{model_training_summary} in Appendix~\secref{app_model_experiments}.

\subsection{AV Dataset to \coda{} Adaptation}
\seclabel{av_adaptation}
\begin{table}[htbp]
\footnotesize
\centering

\renewcommand{\arraystretch}{1.25} 

\begin{tabular}{c|cc|cc|cc|cc}
    \cellcolor{White}{\color{black}\diagbox{\textbf{PT}}{\textbf{DA}}}&
    \multicolumn{2}{c|}{\cellcolor{White}{\color{black}\textbf{Direct}}} &
    \multicolumn{2}{c|}{\cellcolor{White}{\color{black}\textbf{ST}}} &
    \multicolumn{2}{c|}{\cellcolor{White}{\color{black}\textbf{FT}}} &
    \multicolumn{2}{c}{\cellcolor{White}{\color{black}\textbf{ST + FT}}}\\
    \hline
    nuScenes~\cite{nuscenes} & 21.30 & 15.53 & 14.07 & 10.76 & 91.39 & 90.16 & 92.38 & 91.02\\
    Waymo~\cite{Waymo} & 46.20 & 43.11 & 38.27 & 34.36 & \bestbev{93.12} & \besttred{92.07} & 92.36 & 91.18\\
    \coda{} & \textbf{92.08} & \textbf{91.11} &  \multicolumn{2}{c}{\cellcolor{White}{\color{black}\textbf{-}}} &  \multicolumn{2}{c}{\cellcolor{White}{\color{black}\textbf{-}}} & \multicolumn{2}{c}{\cellcolor{White}{\color{black}\textbf{-}}}\\
\end{tabular}

\caption{Evaluation of PV-RCNN pretrained (\textbf{PT}) on AV Datasets and evaluated on the \coda{} test split after undergoing different domain adaptation (\textbf{DA}) methods. 
DA methods include: 1) \textbf{Direct} --- train on the source dataset and evaluate directly on \coda{}; 2) \textit{ST3D++} \textbf{ST}~\cite{yang2021st3d}--- for unsupervised adaptation;  3) Fine-tuning (\textbf{FT})--- with \coda{} after pre-training on the source dataset; and 4) Both ST3D++ and Fine-tuning (\textbf{ST + FT}). The results demonstrate that even state-of-the-art unsupervised domain adaptation methods for 3D object detectors are not competitive with approaches that use domain-specific training labels. All models are evaluated on the medium test split of \coda{}.}

\tablabel{da_analysis}
\end{table}

We apply several domain adaptation strategies to evaluate 3D object detector performance on \coda{} with and without domain-specific labels. In our experiment setup, we choose the object class list to be car, pedestrian, and cyclist so that it is consistent with the standard class list evaluated for nuScenes and Waymo. We perform the standard 3D data augmentation techniques (scaling, rotation, flipping) and align the point cloud ground plane heights for our experiments.

\textbf{Direct Transfer (Direct)}. In this experiment, the campus dataset is not accessible and the pre-trained model is evaluated directly on the test split. This is our baseline for the expected performance when deploying on a campus scale without \coda{}.

\textbf{ST3D++ (ST)}. In this scenario, the campus dataset is accessible but the ground truth labels are not available. This is typical for robot deployments as domain-specific raw sensor data is readily available. We used ST3D++~\cite{yang2021st3d} to adapt to the campus domain as the authors demonstrated state-of-the-art unsupervised domain adaptation performance improvement between different AV datasets when using their method. 

We perform a coarse hyperparameter tuning sweep across the positive and negative thresholds for each object class and use the same model weights for each class. After performing the self-training process for 25 epochs, we evaluate the highest-performing epoch directly on \coda{}. We present the highest performing models in \tabref{model_training_summary} and include the full experiment list in Appendix \secref{app_model_experiments}, \tabref{av_to_coda_training_summary}. This is the best performance we can achieve without labels when deploying on a campus scale with raw sensor data available.

\textbf{Domain Specific Finetuning (FT)}. We assume that domain-specific ground truth labels are available. We pre-train the model backbone on nuScenes or Waymo before fine-tuning on the train split of \coda{}, hypothesizing that learning features on other datasets benefit domain-specific performance.

We pre-train PVRCNN from scratch on each AV dataset for 30 epochs and evaluate the model on the \coda{} test split. After pre-training, we freeze the encoder and backbone weights and randomly initialize the detection, classification, and dense heads. We finetune the heads for 25 epochs, unfreeze the encoder and backbone weights, and train the entire model for another 25 epochs. Our experiments~\ref{tab:model_training_summary} show that a learning rate of 0.01 and the Adam 1cycle optimizer~\cite{SmithCoRR2020} provide the best empirical performance.

\textbf{ST3D++ with Domain Specific Finetuning (ST + FT)}. This experiment combines the ST and FT methods described earlier. We follow the same procedure for self-training using the ST method with the same hyperparameters. After self-training, we apply the training procedure in FT. For this approach, we find that a learning rate of 0.01 and the Adam 1cycle optimizer provides the best empirical performance. 

\textbf{Domain Adaptation Discussion}. \tabref{da_analysis} demonstrates a significant performance gap between unsupervised domain adaptation and fully supervised methods. The highest-performing pre-trained model is about 40 percent lower than the same model trained from scratch on \coda{}. This is expected due to the large domain and sensor-specific differences described earlier. For ST, the model pre-trained on AV datasets decreases in performance after self-training on \coda{}. These results are consistent with findings from the ONCE~\cite{once} AV dataset. They show that performing unsupervised domain adaptation with ST3D from nuScenes to ONCE decreases performance and hypothesize that this is due to differences in LiDAR beam resolution. We believe that differences in sensor viewpoint and resolution cause ST3D to produce poor quality pseudo labels on \coda{} and support this hypothesis in~\secref{sensor_resolution}.

Our experiments show that pre-training on Waymo improves performance on 3D bounding box and BEV tasks by about 1-2 percent compared to training from scratch. However, performance does not consistently improve between FT and ST+FT adaptation techniques. We speculate that when trained to performance saturation, FT improvements dominate the effects of ST pre-training. We conclude from these studies that downstream tasks like 3D object detection benefit from better initial 3D representations. In addition, we hope that our empirical analysis of methods like ST3D spurs future work on how to continue improving self-training methods between domains with significant changes in sensor resolution, viewpoint, and geometric features.

\subsection{Impact of Sensor Resolution Differences on Object Detection Performance}
\seclabel{sensor_resolution}

\begin{table}[ht]
    \centering
    \renewcommand{\arraystretch}{1.25} 
    \begin{tabular}{c|*{4}{cc}}
        {\diagbox{\textbf{Train}}{\textbf{Test}}} &
        \multicolumn{2}{c}{\cellcolor{White}{\color{black}\textbf{\coda{}-16}}} &
        \multicolumn{2}{c}{\cellcolor{White}{\color{black}\textbf{\coda{}-32}}} &
        \multicolumn{2}{c}{\cellcolor{White}{\color{black}\textbf{\coda{}-64}}} &
        \multicolumn{2}{c}{\cellcolor{White}{\color{black}\textbf{\coda{}-128}}}\\
        \hline
        \coda{}-16 & \bestbev{75.15} & \besttred{73.29} & 64.99 & 63.24 & 49.17 & 47.36 & 21.93 & 18.94\\
        \coda{}-32 & 50.79 & 47.95 & \bestbev{78.30} & \besttred{76.90} & 70.49 & 69.37 & 59.95 & 56.59\\
        \coda{}-64 & 21.10 & 22.05 & 67.27 & 64.77 & \bestbev{86.20} & \besttred{84.48} & 77.63 & 77.53\\
        \coda{}-128 & 12.58 & 12.16 & 48.05 & 45.76 & 76.51 & 75.38 & \bestbev{92.61} & \besttred{91.34}\\
    \end{tabular}
    
    \caption{Evaluating the impact of point cloud resolution differences between the source and target domain on 3D object detector performance. All experiments are conducted with a PV-RCNN detector first pre-trained on Waymo using the pedestrian, car, and cyclist classes. We fine-tune the pre-trained model on \coda{} downsampled to 16, 32, 64, and 128 vertical channels (\coda{}-\#channels) for 50 epochs. We then evaluate the model performance on different point cloud resolutions for the pedestrian, car, and cyclist classes using the same evaluation metric as \tabref{base_model_selection}. All models are evaluated on the medium test split of \coda{}. }
    
\tablabel{lidar_resolution_analysis}
\end{table}

While most robots benefit from having high-quality object detections, their wide range of sensor setups presents a challenge for object detectors. Therefore, it is important to understand how object detection performance is affected by sensor resolution differences between the train and test domains.

For our experiments, we train PV-RCNN from scratch on 20\% of the Waymo train dataset and fine-tune the model on the \coda{} medium train split at four LiDAR resolutions (16, 32, 64, 128) on the car, pedestrian, and cyclist classes for 30 epochs or until performance saturates. Our LiDAR is originally 128 channels, so we subsample the original point cloud to obtain the lower resolutions. For fine-tuning, we follow the same two-stage process used in the prior experiments: train a randomly initialized detection head for 15 epochs while keeping the model backbone frozen and then train the full model for another 30 epochs. After training, we evaluate the model directly on the \coda{} medium test split at all four LiDAR resolutions.

\textbf{Sensor Resolution Discussion}. \tabref{lidar_resolution_analysis} shows that 3D object detectors trained on a specific sensor resolution perform best on the same sensor resolution during test time. Furthermore, the larger the resolution difference between the train and test domains, the more performance is affected. This vindicates our hypothesis that large differences in LiDAR resolutions negatively affect object detection performance. Thus, we release pre-trained models on all classes in \coda{} for the (16, 32, 64, and 128) channel LiDAR resolutions and encourage users to select the pre-trained model that is most similar to the target dataset's resolution. \tabref{lidar_resolution_analysis_allclass} in Appendix~\secref{app_model_analysis} reproduces this experiment for all classes in \coda{}.

\subsection{JRDB Adaptation}
\seclabel{jrdb_adaptation}
\begin{table}[ht]
    \footnotesize
    \centering
    
    \renewcommand{\arraystretch}{1.35} 
    \begin{tabular}{c|ccc|ccc}
        \multirow{2}{*}{\diagbox[height=2\line]{\textbf{Train}}{\textbf{Test}}} & \multicolumn{3}{c|}{\textbf{JRDB (15m)} } & \multicolumn{3}{c}{ \textbf{JRDB (25m)} } \\
        & \textbf{Prec.} & \textbf{Rec.} & \textbf{F1} & \textbf{Prec.} & \textbf{Rec.} & \textbf{F1} \\
        \hline
        Waymo & 55.39 & 18.70 & 27.96 & 52.76 & 17.19 & 25.94\\
        \coda{} & 60.29\textsuperscript{\textcolor{blue}{+4.90}} & 25.32\textsuperscript{\textcolor{blue}{+6.62}} & 35.66\textsuperscript{\textcolor{blue}{+7.7}} & 57.38\textsuperscript{\textcolor{blue}{+4.62}} & 25.31\textsuperscript{\textcolor{blue}{+8.12}} & 35.13\textsuperscript{\textcolor{blue}{+9.19}} \\
        \hline
        JRDB & 65.64 & 27.14 & 38.39 & 64.15 & 27.15 & 38.15\\
    \end{tabular}
    
    \caption{Cross-dataset 3D object detection performance comparison on JRDB~\cite{JRDB} after training on 
    \coda{} and Waymo~\cite{Waymo}. We train a PV-RCNN detector on only pedestrians for all datasets. We evaluate the average precision, recall, and F1 score for objects within 15 meters (15m) and 25 meters (25m) of the ego vehicle. We report the performance difference between the AV and \coda{} models in red and blue superscripts.}
    
    \tablabel{jrdb_zeroshot_analysis}
\end{table}

\begin{table*}[ht]
\footnotesize
\centering

\renewcommand{\arraystretch}{1.25} 


\begin{tabular}{cc|c|c|c|c|c|c|c|c|c|c|c|c|c|c|c|c|c|c|c|c|c|c|c}
    \cellcolor{white}{\color{black}\textbf{Approach}} &
    \rotatebox{90}{\cellcolor{white}{\color{black}\textbf{mIoU}}} &
    \rotatebox{90}{\cellcolor{concreteColor}{\color{white}\textbf{Concrete}}} &
    \rotatebox{90}{\cellcolor{grassColor}{\color{white}\textbf{Grass}}} &
    \rotatebox{90}{\cellcolor{rocksColor}{\color{white}\textbf{Rocks}}} &
    \rotatebox{90}{\cellcolor{speedwayBricksColor}{\color{white}\textbf{Speedway Bricks}}} &
    \rotatebox{90}{\cellcolor{redBricksColor}{\color{white}\textbf{Red Bricks}}} &
    \rotatebox{90}{\cellcolor{pebblePavementColor}{\color{white}\textbf{Pebble Pavement}}} &
    \rotatebox{90}{\cellcolor{lightMarbleTilingColor}{\color{white}\textbf{Light Marble Tiling}}} &
    \rotatebox{90}{\cellcolor{darkMarbleTilingColor}{\color{white}\textbf{Dark Marble Tiling}}} &
    \rotatebox{90}{\cellcolor{dirtPathsColor}{\color{white}\textbf{Dirt Paths}}} &
    \rotatebox{90}{\cellcolor{roadPavementColor}{\color{white}\textbf{Road Pavement}}} &
    \rotatebox{90}{\cellcolor{shortVegetationColor}{\color{white}\textbf{Short Vegetation}}} &
    \rotatebox{90}{\cellcolor{porcelainTileColor}{\color{white}\textbf{Porcelain Tile}}} &
    \rotatebox{90}{\cellcolor{metalGratesColor}{\color{white}\textbf{Metal Grates}}} &
    \rotatebox{90}{\cellcolor{blondMarbleTilingColor}{\color{white}\textbf{Blond Marble Tiling}}} &
    \rotatebox{90}{\cellcolor{woodPanelColor}{\color{white}\textbf{Wood Panel}}} &
    \rotatebox{90}{\cellcolor{patternedTileColor}{\color{white}\textbf{Patterned Tile}}} &
    \rotatebox{90}{\cellcolor{carpetColor}{\color{white}\textbf{Carpet}}} &
    \rotatebox{90}{\cellcolor{crosswalkColor}{\color{white}\textbf{Crosswalk}}} &
    \rotatebox{90}{\cellcolor{domeMatColor}{\color{white}\textbf{Dome Mat}}} &
    \rotatebox{90}{\cellcolor{stairsColor}{\color{white}\textbf{Stairs}}} &
    \rotatebox{90}{\cellcolor{doorMatColor}{\color{white}\textbf{Door Mat}}} &
    \rotatebox{90}{\cellcolor{thresholdColor}{\color{white}\textbf{Threshold}}} &
    \rotatebox{90}{\cellcolor{metalFloorColor}{\color{white}\textbf{Metal Floor}}}\\
    \hline
    Cylinder3D~\cite{Cylinder3D} & 49.9 & \textbf{68.7} & \textbf{54.7} & 0.2 & \textbf{69.7} & 57.4 & 38.2 & 48.2 & 45.3 & \textbf{70.9} & \textbf{80.6} & \textbf{81.8} & 81.2 & \textbf{39.2} & 7.9 & \textbf{78.8} & 3.5 & 79.0 & \textbf{52.8} & \textbf{24.9} & \textbf{93.9} & 10.7 & \textbf{5.3} & \textbf{59.6}\\
    2DPass~\cite{yan20222dpass} & \textbf{51.5} & 51.2 & 36.3 & \textbf{68.1} & 67.0 & \textbf{63.6} & \textbf{39.4} & \textbf{62.9} & \textbf{68.6} & 61.4 & 64.4 & 70.5 & \textbf{83.2} & 13.2 & \textbf{34.2} & 73.9 & \textbf{81.5} & \textbf{84.1} & 34.5 & 0.1 & 87.1 & \textbf{29.0} & 0.0 & 25.9\\
\end{tabular}

\caption{Evaluation of two 3D semantic segmentation models on the full \coda{} test split. We report mean intersection over union and accuracy for each semantic class. Bold numbers indicate the highest-performing method for each category.} 

\tablabel{semseg_benchmarks}
\end{table*}

Aside from sensor variations, viewpoint and scene differences between train and test domains also present a challenge for LiDAR-based object detectors. To understand the impact of these differences, we evaluate the performance of 3D object detectors trained on \coda{} and AV datasets on JRDB, a large-scale urban robot dataset with LiDAR point clouds and 3D bounding box annotations.

For our experiments, we train three PV-RCNN models from scratch on 20\% of the Waymo train split, full \coda{} train split, and full JRDB train split for 30 epochs or until performance saturates. For consistency, all models are only trained on pedestrians and evaluated on the proposed JRDB validation split using their 3D detection benchmark metrics (average precision, recall, and F1 score). We repeat our evaluation on two variations of the validation split: one containing ground truth annotations exclusively within 15 meters of the ego vehicle and the other within 25 meters of the ego vehicle.

\textbf{JRDB Performance Discussion.} \tabref{jrdb_zeroshot_analysis} shows that \coda{} models consistently outperform Waymo models in all metrics for both the 15m and 25m range. Furthermore, pretraining on \coda{} offers similar performance to training with labels on JRDB, corroborating our claim that pre-trained \coda{} models generalize to other urban settings. We believe this can be explained by \coda{}'s similarity to JRDB in terms of sensor resolution, viewpoint, and scene diversity. By utilizing prior knowledge of similar environments in JRDB, \coda{} models are more robust to point cloud sparsity than Waymo models. \figref{jrdbtpfpfn} in \secref{jrdb_adaptation} vindicates our claim with several examples where \coda{} models detect sparse pedestrians that Waymo models miss. 

To assess how variations in sensor resolution affect model performance across datasets, we evaluated models trained on different resolutions of \coda{} on JRDB in \tabref{jrdb_hyperparameter_summary} in Appendix~\secref{jrdb_adaptation}. Our findings indicate that detection performance decreases as the sensor resolution difference increases between the train and test datasets. This aligns with the insights we presented in \secref{sensor_resolution}, demonstrating that cross-dataset performance is maximized when the train and test resolutions closely match. Thus, we recommend that users select the pre-trained model that is most similar to their target dataset's resolution for optimal performance. Our findings should motivate future work to leverage scene context and develop density invariant models to improve 3D object detection performance.

\section{Benchmarks}

In this section, we define the 3D object detection and 3D semantic segmentation for this dataset. We plan on adding additional tasks in the future for robot perception and planning, such as long-term SLAM, cross-domain information retrieval, and preference-aware navigation.

\subsection{3D Object Detection}

The 3D object detection task involves predicting 7 degrees of freedom boxes for all object classes. We use the 3D object detection metric proposed in the KITTI Vision Benchmark Suite. For the car, pedestrian, and cyclist classes, we require a minimum bounding box overlap of 70\%, 50\%, and 50\% to determine if detection is correct. For all other object classes, we use a minimum overlap of 50\% with the ground truth bounding box. All methods are limited to using up to 10 prior LiDAR frames for predictions. All sensor modalities and pseudo-ground truth poses can be used, and we will evaluate all predictions on the 3D point cloud annotations.

\subsection{3D Semantic Segmentation}

For the 3D semantic segmentation benchmark, we use the same evaluation metric proposed in SemanticKITTI~\cite{SemanticKitti}. This is the mean intersection-over-union (mIoU) metric~\cite{PascalVOC} over all classes. All sensor modalities can be used, but we will evaluate all predictions using the 3D point cloud annotations. \tabref{semseg_benchmarks} benchmarks Cylinder3D~\cite{Cylinder3D} and 2DPass~\cite{yan20222dpass}, two state-of-the-art LiDAR only and LiDAR camera approaches respectively. For our benchmarks, we train both models from scratch for 30 epochs or until performance saturates and take the highest-performing model.

\section{Conclusion and Future Work}

In this work, we presented the UT Campus Object Dataset (\coda{}), a multi-modal dataset that contains greater object and scene-level annotation diversity than any other similar existing dataset. \coda{} contains 1.3 million human-annotated 3D bounding boxes and 5000 semantic segmentation annotations over 8.5 hours of data collected from the perspective of a mobile robot across UT campus. We publicly release \coda{} on the~\href{https://doi.org/10.18738/T8/BBOQMV}{Texas Data Repository}~\cite{coda2023tdr}, \href{https://github.com/ut-amrl/coda-models}{pre-trained models} for various LiDAR resolutions (16, 32, 64, 128 channels), and \href{https://github.com/ut-amrl/coda-devkit}{dataset development package}.

We conducted extensive experiments to select a high-performing model architecture for urban environments. We demonstrated a performance gap for 3D object detectors in urban environments by comparing the performance on \coda{}'s test split after training on \coda{} versus AV datasets. We empirically demonstrated that 3D object detection performance is significantly affected by differences in LiDAR sensor resolution during test time.  Finally, we conducted various ablation studies to show that pre-training on \coda{} instead of AV datasets improves cross-dataset object detection performance on existing urban robotics datasets. This constitutes motivation for future work to improve 3D object detector invariance to point cloud density and highlights the importance of selecting a pre-trained model that closely resembles the target domain during robot deployments. We expect that this work will spur future research toward learning sensor-invariant 3D feature representations, object-centric localization approaches, and terrain-aware navigation planners. In the future, we plan on releasing additional benchmarks on \coda{} to facilitate fair comparison for methods in these research areas.

\section{Acknowledgement}

This work was conducted with the Autonomous Mobile Robotics Laboratory (AMRL) at UT Austin. This project is partially supported by NSF Awards CAREER-2046955 and IIS-1954778. We would like to thank Roberto Mart\'in-Mart\'in, Zhangyang "Atlas" Wang, and Philipp Kr\"ahenb\"uhl for their fruitful comments, suggestions, and inspiration. The authors acknowledge the Texas Advanced Computing Center (TACC)\footnote{TACC website: \url{http://www.tacc.utexas.edu}} at The University of Texas at Austin for providing HPC resources that have contributed to the research results reported within this paper. URL: http://www.tacc.utexas.edu. The \coda{} study is covered under the University of Texas at Austin IRB Number STUDY00003493.

\bibliographystyle{IEEEtran}
\bibliography{refs}

\clearpage
\section{Appendix}
\seclabel{appendix}

We organize the appendix into the following sections: extended model analysis (\secref{app_model_analysis}), \coda{} organization structure (\secref{app_organization_structure}), model training experiments (\secref{app_model_experiments}), annotation ontology (\secref{app_annotation_onotology}), qualitative 3D object detection results (\secref{app_qualitative_results}), and ground truth annotation visualizations (\secref{app_gt_visualizations}).

\subsection{Extended Model Analysis}
\seclabel{app_model_analysis}

\tabref{lidar_resolution_analysis_allclass} repeats the analysis in \tabref{lidar_resolution_analysis} for all annotated classes in \coda{}. We conduct these experiments using the same experimental setup and conclude that our findings in \secref{sensor_resolution} hold for all classes. As such, users should select the pre-trained model with the closest LiDAR resolution to their target domain for optimal performance.

\begin{table}[ht]
    \centering
    \renewcommand{\arraystretch}{1.25} 
    \begin{tabular}{c|*{4}{cc}}
        {\diagbox{\textbf{Train}}{\textbf{Test}}} &
        \multicolumn{2}{c}{\cellcolor{White}{\color{black}\textbf{\coda{}-16}}} &
        \multicolumn{2}{c}{\cellcolor{White}{\color{black}\textbf{\coda{}-32}}} &
        \multicolumn{2}{c}{\cellcolor{White}{\color{black}\textbf{\coda{}-64}}} &
        \multicolumn{2}{c}{\cellcolor{White}{\color{black}\textbf{\coda{}-128}}}\\
        \hline
        \coda{}-16 & \bestbev{23.36} & \besttred{21.15} & 19.39 & 17.03 & 9.65 & 8.37 & 2.99 & 2.38\\
        \coda{}-32 & 14.52 & 12.23 & \bestbev{26.19} & \besttred{23.86} & 23.38 & 21.28 & 17.10 & 15.08\\
        \coda{}-64 & 5.70 & 4.93 & 20.49 & 18.29 & \bestbev{27.38} & \besttred{25.51} & 25.13 & 23.30\\
        \coda{}-128 & 3.14 & 2.53 & 12.85 & 11.31 & 25.40 & 23.34 & \bestbev{28.18} & \besttred{26.14}\\
    \end{tabular}
    
    \caption{Evaluating the impact of point cloud resolution differences between the source and target domain on 3D object detector performance. All experiments are conducted with a PV-RCNN~\cite{pvrcnn} detector first pre-trained on Waymo~\cite{Waymo} using the pedestrian, car, and cyclist classes. We fine-tune the pre-trained model on the full \coda{} train split downsampled to 16, 32, 64, and 128 vertical channels (\coda{}-\#channels) for 30 epochs. Models are finetuned on all classes in \coda{} and evaluated using the same evaluation metric as \tabref{base_model_selection}.}
    \tablabel{lidar_resolution_analysis_allclass}
\end{table}

\subsection{\coda{} Organization Structure}
\seclabel{app_organization_structure}
\begin{figure}[ht]
    \centering
    \includegraphics[width=0.9\linewidth]{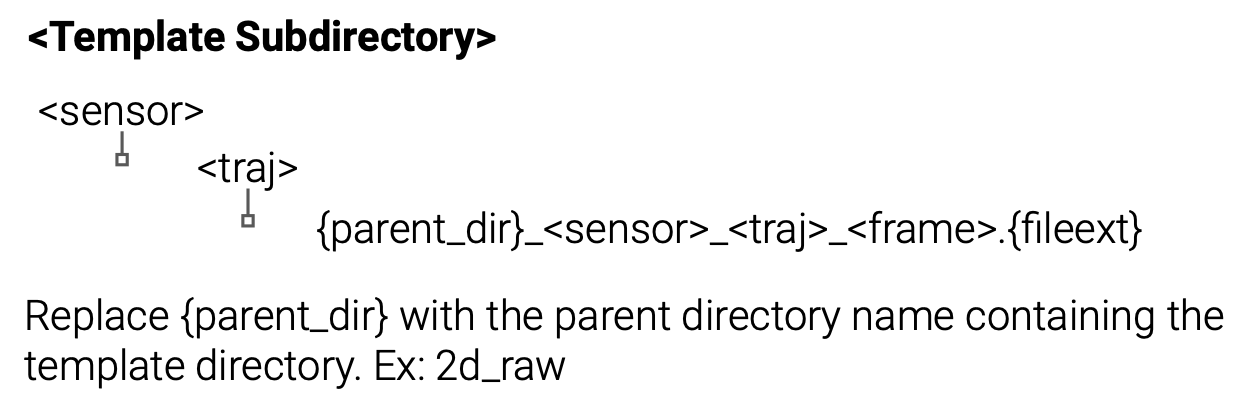}
    \caption{Full 3D bounding box object list. Bolding represents the topological category for the semantic classes below.}
    \figlabel{CODaTemplateSubdirectory}
\end{figure}

\begin{figure}[htbp]
    \centering
    \includegraphics[width=0.6\columnwidth]{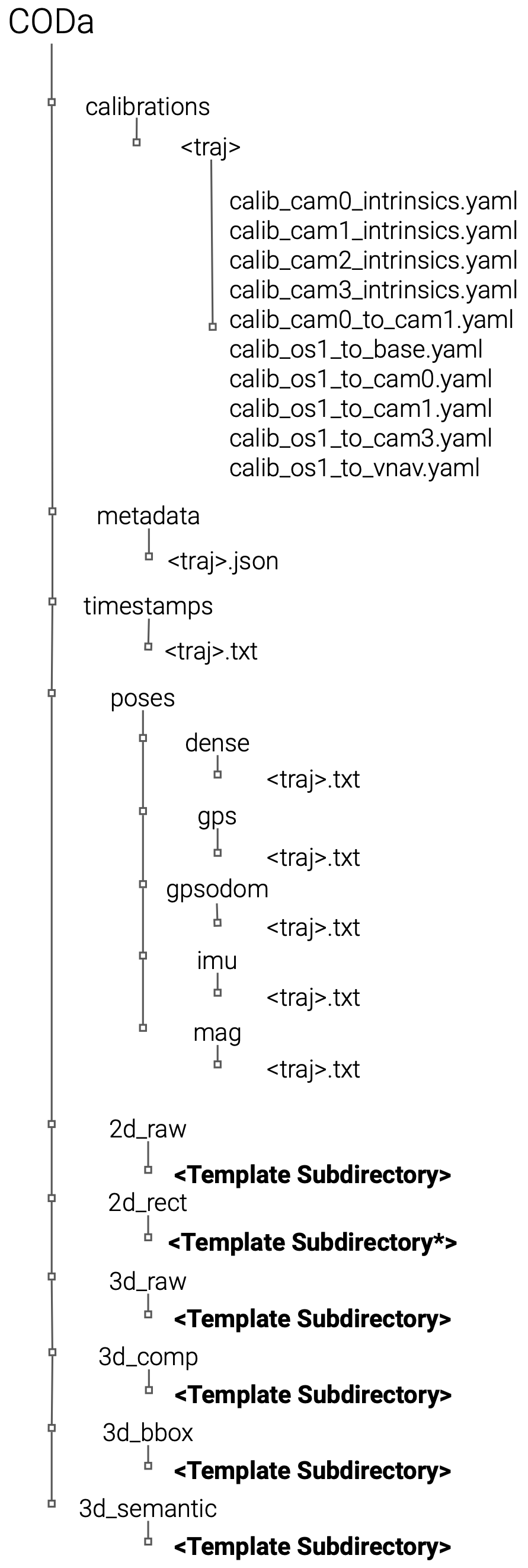}
    \caption{Directory structure of \coda{}. All sensor and annotation directories use the subdirectory structure found in \figref{CODaTemplateSubdirectory}. The full dataset contains point cloud, image, inertial, sensor calibrations, timestamps, and annotations. More details can be found on the data report~\cite{coda2023tdr}. }
    \figlabel{CODaDirectory}
\end{figure}

We describe the organization structure of \coda{} at a high level and refer the reader to the data report~\cite{coda2023tdr} for a detailed breakdown of the file contents. The primary non-sensor subdirectories in \coda{} contain sensor calibrations, metadata, timestamps, and poses. We provide these files for each sequence. The remaining directories contain 2D/3D sensor data and annotations. Each of these directories has an identical subdirectory structure to \figref{CODaTemplateSubdirectory}. The metadata files contain information on each sequence, such as the list of semantic objects present and the dataset splits by task.

\begin{table*}[htbp]
    \footnotesize
    \centering
    
    \renewcommand{\arraystretch}{1.1} 
    
    \normalsize
    \begin{tabular}{cc|cc|ccc|ccc|cc}
        \cellcolor{White}{\color{black}\textbf{Backbone}}&
        \cellcolor{White}{\color{black}\textbf{Head}}&
        \cellcolor{White}{\color{black}\textbf{VS}}&
        \cellcolor{White}{\color{black}\textbf{P/V}}&
        \cellcolor{White}{\color{black}\textbf{LR}}&
        \cellcolor{White}{\color{black}\textbf{OPT}}&
        \cellcolor{White}{\color{black}\textbf{Epochs}}&
        \cellcolor{White}{\color{black}\textbf{Train}}&
        \cellcolor{White}{\color{black}\textbf{FT}}&
        \cellcolor{White}{\color{black}\textbf{Eval}}&
        \cellcolor{White}{\color{black}\textbf{AP$_{BEV}$}}&
        \cellcolor{White}{\color{black}\textbf{AP$_{3D}$}}\\
        \hline
        PointPillars & AnchorMulti & 0.1 & 5 & 3e-3 & adam1cycle & 30 & Waymo & - & Waymo & 21.86 & 16.78\\
        PointPillars & AnchorSingle & 0.32 & 20 & 3e-3 & adam1cycle & 30 & Waymo & - & Waymo & \textbf{55.11} & \textbf{47.55}\\
        PointPillars & AnchorMulti & 0.1 & 5 & 1e-3 & adam1cycle & 30 & nuScenes & - & nuScenes & \textbf{28.42} & \textbf{17.94}\\
        PointPillars & AnchorMulti & 0.2 & 5 & 1e-3 & adam1cycle & 30 & nuScenes & - & nuScenes & 24.14 & 14.28\\
        PointPillars & AnchorMulti & 0.1 & 5 & 3e-3 & adam1cycle & 80 & KITTI & - & KITTI & 27.99 & 25.03\\
        PointPillars & AnchorSingle & 0.1 & 32 & 3e-3 & adam1cycle & 80 & KITTI & - & KITTI & \textbf{70.27} &\textbf{63.32}\\
        PointPillars & AnchorSingle & 0.16 & 32 & 3e-3 & adam1cycle & 80 & KITTI & - & KITTI & 70.94 & 64.49\\
        PointPillars & AnchorMulti & 0.1 & 5 & 3e-3 & adam1cycle & 50 & \coda{} & - & \coda{} & 15.96 & 14.75\\
        PointPillars & AnchorSingle & 0.1 & 5 & 3e-3 & adam1cycle & 50 & \coda{} & - & \coda{} & \textbf{49.78} & \textbf{48.86}\\
        \hline
        CenterPoint-Pillar & CenterHead & 0.1 & 5 & 3e-3 & adam1cycle & 50 & Waymo & - & Waymo & 39.38 & 31.67\\
        CenterPoint-Voxel & CenterHead & 0.1 & 5 & 3e-3 & adam1cycle & 30 & Waymo & - & Waymo & \textbf{62.66} & \textbf{54.86}\\
        CenterPoint-Pillar & CenterHead & 0.1 & 10 & 1e-3 & adam1cycle & 50 & nuScenes & - & nuScenes & 21.82 & 12.08\\
        CenterPoint-Voxel & CenterHead & 0.1 & 10 & 3e-3 & adam1cycle & 50 & nuScenes & - & nuScenes & 29.17 & 18.79\\
        CenterPoint-Voxel & CenterHead & 0.075 & 10 & 1e-3 & adam1cycle & 30 & nuScenes & - & nuScenes & \textbf{36.91} & \textbf{23.86}\\
        CenterPoint-Pillar & CenterHead & 0.1 & 5 & 3e-3 & adam1cycle & 80 & KITTI & - & KITTI & \textbf{69.34} & \textbf{63.87}\\
        CenterPoint-Voxel & CenterHead & 0.1 & 5 & 3e-3 & adam1cycle & 80 & KITTI & - & KITTI & 66.83 & 60.15\\
        CenterPoint-Pillar & CenterHead & 0.1 & 5 & 1e-2 & adam1cycle & 50 & \coda{} & - & \coda{} & 61.78 & 52.46\\
        CenterPoint-Voxel & CenterHead & 0.075 & 10 & 3e-3 & adam1cycle & 50 & \coda{} & - & \coda{} & \textbf{82.08} & \textbf{76.92}\\
        \hline
        PVRCNN & AnchorSingle & 0.1 & 5 & 1e-2 & adam1cycle & 50 & Waymo & - & Waymo & 43.60 & 48.41\\
        PVRCNN & CenterHead & 0.1 & 5 & 1e-2 & adam1cycle & 30 & Waymo & - & Waymo & \textbf{62.73} & \textbf{56.40}\\
        PVRCNN & AnchorSingle & 0.1 & 10 & 1e-2 & adam1cycle & 30 & nuScenes & - & nuScenes & 30.89 & 21.74\\
        PVRCNN & CenterHead & 0.1 & 5 & 1e-2 & adam1cycle & 50 & nuScenes & - & nuScenes & \textbf{33.85} & \textbf{25.41}\\
        PVRCNN & AnchorSingle & 0.1 & 5 & 1e-2 & adam1cycle & 80 & KITTI & - & KITTI & \textbf{70.22} & \textbf{65.28}\\
        PVRCNN & CenterHead & 0.1 & 20 & 1e-2 & adam1cycle & 50 & \coda{} & - & \coda{} & \textbf{92.08} & \textbf{91.11}\\
        \hline
    \end{tabular}
    
    \caption{Full model training summary for \tabref{base_model_selection}. We evaluate several 3D object detection architectures on AV Datasets and \coda{}. We separate the table columns by model architecture, data processing parameters, training hyperparameters, dataset, and performance. The column abbreviations are as follows: VS - voxel length and width, P/V - points per voxel, LR - learning rate, OPT - optimizer, FT - finetune dataset, Eval - evaluation dataset. We train the the PointPillars~\cite{lang2019pointpillars}, Centerpoint~\cite{centerpoint}, and PVRCNN~\cite{pvrcnn} detectors on 20\% of Waymo~\cite{Waymo}, all of nuScenes~\cite{nuscenes}, all of KITTI~\cite{KITTIVisionSuite}, and the medium split of \coda{}. We report mean average precision for the car, pedestrian, and cyclist categories in bird's eye view (AP$_{BEV}$) and 3D (AP$_{3D}$) with IOU 0.7, 0.5, and 0.5 respectively. We average the results at the easy, medium, and hard difficulties. Results included in the main paper are bolded.}
    \tablabel{model_training_summary}
    \end{table*}

\subsection{Model Training Experiments}
\seclabel{app_model_experiments}

In this section, we present all model experiments for the object detector architecture selection found in \tabref{base_model_selection}, AV to \coda{} domain adaptation in \tabref{da_analysis}, cross sensor resolution experiments in \tabref{lidar_resolution_analysis}, and adaptation to JRDB in \tabref{jrdb_zeroshot_analysis}. 

\textbf{Architecture Selection Experiments}. Object detection detection performance is highly dependent on the architecture and preprocessing hyperparameters. We limit our evaluation to PointPillars, CenterPoint, and PVRCNN for reasons discussed in \secref{results}. \tabref{model_training_summary} summarizes the model configurations considered. Broadly speaking, we find that performance is highly dependent on the voxel size and points per voxel. While this is not true for all models, we find that a voxel size and points per voxel of 0.1 and 5 often improve performance. We do not perform a hyperparameter search over learning rates and optimizers for AV datasets because the OpenPCDet contributors already provide highly performant learning rates and optimizers. For \coda{}, we find that using the same learning rate as AV datasets offers good performance.

\begin{table*}[htbp]
    \footnotesize
    \centering
    
    \renewcommand{\arraystretch}{1.1} 
    
    \normalsize
    \begin{tabular}{cc|ccc|cc|cc|ccc|cc}
        \cellcolor{White}{\color{black}\textbf{Backbone}}&
        \cellcolor{White}{\color{black}\textbf{Head}}&
        \cellcolor{White}{\color{black}\textbf{DA}}& 
        \cellcolor{White}{\color{black}\textbf{MV}}& 
        \cellcolor{White}{\color{black}\textbf{+ST}}& 
        \cellcolor{White}{\color{black}\textbf{VS}}&
        \cellcolor{White}{\color{black}\textbf{P/V}}&
        \cellcolor{White}{\color{black}\textbf{LR}}&
        \cellcolor{White}{\color{black}\textbf{Epochs}}&
        \cellcolor{White}{\color{black}\textbf{Train}}&
        \cellcolor{White}{\color{black}\textbf{FT}}&
        \cellcolor{White}{\color{black}\textbf{Eval}}&
        \cellcolor{White}{\color{black}\textbf{AP$_{BEV}$}}&
        \cellcolor{White}{\color{black}\textbf{AP$_{3D}$}}\\
        \hline
        PVRCNN & CenterHead & Direct & Y & - & 0.1 & 5 & 1e-2 & 30 & nuScenes & - & \coda{}-md & \textbf{21.30} & \textbf{15.53}\\
        PVRCNN & CenterHead & Direct & Y & - & 0.1 & 5 & 1e-2 & 30 & Waymo & - & \coda{}-md & \textbf{46.20} & \textbf{43.11}\\
        \hline
        PVRCNN & AnchorSingle & ST & Y & 0.4, 0.5, 0.5 & 0.1 & 5 & 1e-3 & 30 & nuScenes & - & \coda{}-sm & 7.65 & 5.39\\
        PVRCNN & AnchorSingle & ST & Y & 0.4, 0.3, 0.3 & 0.1 & 5 & 1e-2 & 30 & nuScenes & - & \coda{}-sm & 16.91 & 15.34\\
        PVRCNN & CenterHead & ST & Y & 0.4, 0.3, 0.3 & 0.1 & 5 & 1e-2 & 30 & nuScenes & - & \coda{}-sm & 23.20 & 19.02\\
        PVRCNN & CenterHead & ST & Y & 0.6, 0.55, 0.55 & 0.1 & 5 & 1e-2& 30 & nuScenes & - & \coda{}-sm & 40.25 & 35.36\\
        PVRCNN & CenterHead & ST & Y & 0.6, 0.55, 0.55 & 0.1 & 5 & 1e-2& 30 & nuScenes & - & \coda{}-md & \textbf{14.07} & \textbf{10.76}\\
        PVRCNN & CenterHead & ST & Y & 0.6, 0.55, 0.55 & 0.1 & 5 & 1e-2 & 30 & Waymo & - & \coda{}-md & 41.55 & 37.32\\
        PVRCNN & CenterHead & ST & Y & 0.4, 0.3, 0.3 & 0.1 & 5 & 1.5e-3 & 30 & Waymo & - & \coda{}-md & 35.70 & 32.51\\
        PVRCNN & CenterHead & ST & Y & 0.6, 0.55, 0.55 & 0.1 & 5 & 1.5e-3 & 30 & Waymo & - & \coda{}-md & \textbf{38.27} & \textbf{34.36}\\
        \hline
        PVRCNN & CenterHead & FT & Y & 0.6, 0.55, 0.55 & 0.1 & 5 & 1e-2 & 30 & nuScenes & \coda{}-md & \coda{}-md & \textbf{91.39} & \textbf{90.16}\\
        PVRCNN & CenterHead & FT & Y & 0.6, 0.55, 0.55 & 0.1 & 5 & 1e-4 & 30 & Waymo & \coda{}-md & \coda{}-md & 52.22 & 48.07\\
        PVRCNN & CenterHead & FT & Y & 0.6, 0.55, 0.55 & 0.1 & 5 & 1e-3 & 30 & Waymo & \coda{}-md & \coda{}-md & 57.51 & 56.01\\
        PVRCNN & CenterHead & FT & Y & 0.6, 0.55, 0.55 & 0.1 & 5 & 1e-2 & 30 & Waymo & \coda{}-md & \coda{}-md & \textbf{93.12} & \textbf{92.07}\\
        \hline
        PVRCNN & CenterHead & ST+FT & Y & 0.6, 0.55, 0.55 & 0.1 & 5 & 1.5e-3 & 15 & nuScenes & \coda{}-md & \coda{}-md & \textbf{91.87} & \textbf{89.89}\\
        PVRCNN & CenterHead & ST+FT & Y & 0.6, 0.55, 0.55 & 0.1 & 5 & 1e-2 & 30 & Waymo & \coda{}-md & \coda{}-md & \textbf{92.36} & \textbf{91.18}\\
        
    \end{tabular}
    \caption{Full model training summary for \tabref{da_analysis}. Results reported in \tabref{da_analysis} are bolded. We train the PVRCNN~\cite{pvrcnn} detector on 20\% of Waymo~\cite{Waymo} and all of nuScenes~\cite{nuscenes}. \coda{} is divided into small (25\%) and medium (50\%) splits of the full dataset. The column abbreviations are as follows: DA - domain adaptation method (refer to \secref{av_adaptation} for definitions of these methods), MV - memory voting hyperparameter, +ST - positive score threshold hyperparameter, VS - voxel size, P/V - points per voxel LR - learning rate, FT - finetuning dataset, Eval - evaluation dataset. Results included in the main paper are bolded.}
    \tablabel{av_to_coda_training_summary}
\end{table*}

\begin{table}[htbp]
    \footnotesize
    \centering
    
    \renewcommand{\arraystretch}{1.1} 
    
    \normalsize
    \begin{tabular}{ccc|cc|ccc}
        \cellcolor{White}{\color{black}\textbf{Range}}&
        \cellcolor{White}{\color{black}\textbf{CT}}&
        \cellcolor{White}{\color{black}\textbf{Augs}}&
        \cellcolor{White}{\color{black}\textbf{Train}}&
        \cellcolor{White}{\color{black}\textbf{Test}}&
        \cellcolor{White}{\color{black}\textbf{AP}}&
        \cellcolor{White}{\color{black}\textbf{AR}}&
        \cellcolor{White}{\color{black}\textbf{AF}}\\
        \hline
        15m & 0.1 & Def. & Waymo & JRDB & \textbf{55.39} & \textbf{18.7} & \textbf{27.96}\\
        25m & 0.1 & Def. & Waymo & JRDB & \textbf{52.76} & \textbf{17.19} & \textbf{25.94}\\
        \hline
        15m & 0.1 & Def. & JRDB & JRDB & \textbf{65.64} & \textbf{27.13} & \textbf{38.39}\\
        25m & 0.1 & Def. & JRDB & JRDB & \textbf{64.15} & \textbf{27.15} & \textbf{38.15}\\
        \hline
        15m & 0.1 & All & \coda{}-32 & JRDB & \textbf{60.29} & \textbf{25.32} & \textbf{35.66}\\
        15m & 0.1 & Def. & \coda{}-64 & JRDB & 57.30 & 23.56 & 33.39\\
        25m & 0.1 & Def. & \coda{}-16 & JRDB & 52.22 & 23.57 & 32.49\\
        25m & 0.1 & All & \coda{}-16 & JRDB & 54.32 & 25.32 & 34.54\\
        25m & 0.1 & Def. & \coda{}-32 & JRDB & 55.42 & 21.89 & 31.39\\
        25m & 0.1 & All & \coda{}-32 & JRDB & \textbf{57.38} & \textbf{25.31} & \textbf{35.13}\\
        25m & 0.1 & Def. & \coda{}-64 & JRDB & 56.49 & 20.26 & 29.83\\
        25m & 0.1 & All & \coda{}-64 & JRDB & 56.27 & 20.25 & 29.78\\
        25m & 0.1 & Def. & \coda{}-128 & JRDB & 54.31 & 18.69 & 27.81\\
    \end{tabular}
    \caption{Full model training summary for \tabref{jrdb_zeroshot_analysis}. Results reported in \tabref{jrdb_zeroshot_analysis} are bolded. We train the PVRCNN~\cite{pvrcnn} detector on 20\% of Waymo~\cite{Waymo} and all of \coda{}. For JRDB, we separate the train/validation splits for JRDB~\cite{JRDB} into two datasets: one excluding labels farther than 15 meters from the ego-vehicle (15m) and the other with all labels within 25 meters (25m). We indicate the data augmentations used by default (Def.) and all (All). The confidence threshold (CT) is the threshold required to be considered a detection. We report the mean average precision (AP), recall (AR), and F1 score (AF) using an IOU of 0.3 on the JRDB validation split. Results included in the main paper are bolded.}
    \tablabel{jrdb_hyperparameter_summary}
\end{table}

\textbf{AV to \coda{} Experiments}. We conduct the following experiments to optimize unsupervised and finetuning performance on \coda{} after pretraining on AV datasets. We tuned the hyperparameters for the ST3D (ST) and finetuning (FT) domain adaptation strategies. \secref{results} describes the experimental setup in detail. \tabref{av_to_coda_training_summary} demonstrates that a high positive score threshold is beneficial for unsupervised domain adaptation. We speculate this is because a high positive threshold filters out more low-confidence detections, reducing the amount of erroneous pseudo-ground truth detections from ST3D. We perform a coarse hyperparameter sweep over the learning rate on the medium \coda{} split (\coda{}-md) and find that higher learning rates significantly improve finetuning performance.

\textbf{JRDB Adaptation Experiments}. We conduct the following experiments to quantitatively evaluate cross-dataset object detection performance on JRDB. We use the experiment setup and evaluation metrics described in \secref{results}, training a PVRCNN detector on Waymo, multiple LiDAR resolutions of the full \coda{} split, and JRDB before evaluating on JRDB. We empirically assess data augmentation effects using two sets: the default suite (random world flipping, rotation, and scaling) and the complete suite from previous cross-dataset object detection research~\cite{2020objdetgeneralize} (random object scaling, rotation). \tabref{jrdb_hyperparameter_summary} presents these results, showing that using the full suite of data augmentation techniques benefits cross-dataset performance. Lastly, we quantify how sensor resolution differences affect cross-dataset detection performance in \tabref{jrdb_hyperparameter_summary}, showing that performance is maximized when \coda{}'s resolution matches JRDB's LiDAR resolution (32 channels). 

\begin{figure}[htbp]
    \centering
    \includegraphics[width=\linewidth]{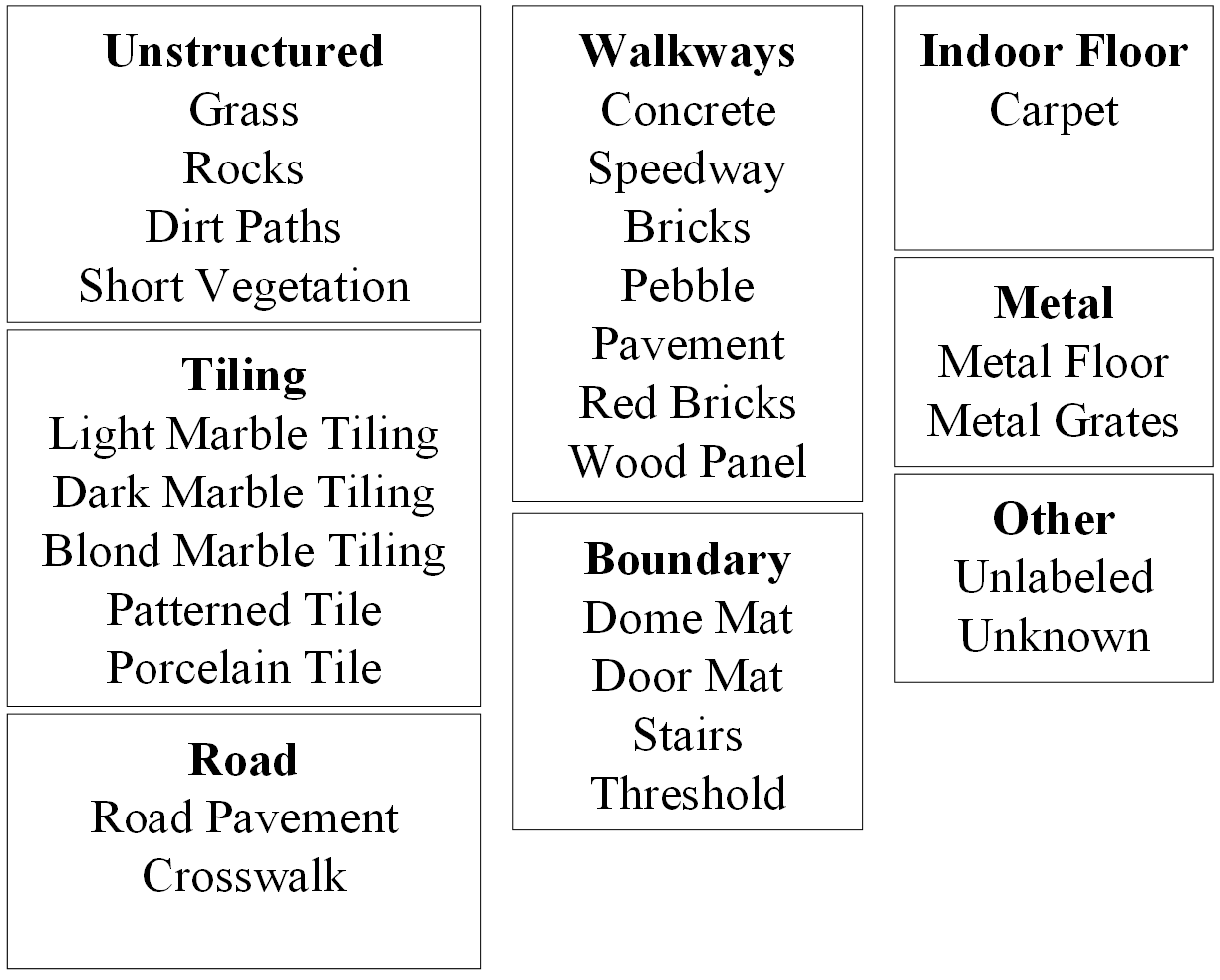}
    \caption{Full 3D semantic segmentation class list. Bolding represents the topological category for the semantic classes below.}
    \figlabel{fullsemseglist}
\end{figure}

\begin{figure*}[htbp]
    \centering
    \includegraphics[width=\linewidth]{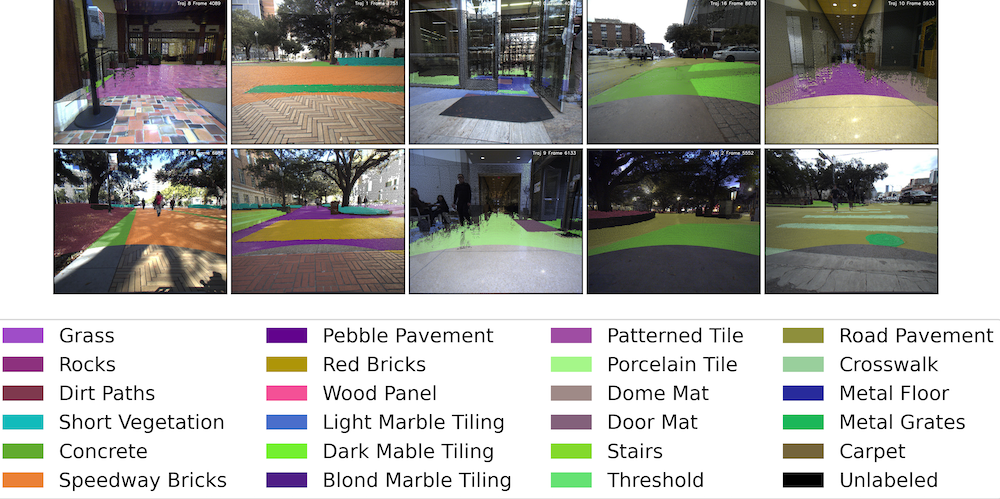}
    \caption{Terrain segmentation colormap for \coda{}. Segmentation labels are verified on the 3D point clouds by human annotators and projected onto 2D images for visualization purposes.}
    \figlabel{semantic2dexamples}
\end{figure*}

\begin{figure}[htbp]
    \centering
    \includegraphics[width=\linewidth]{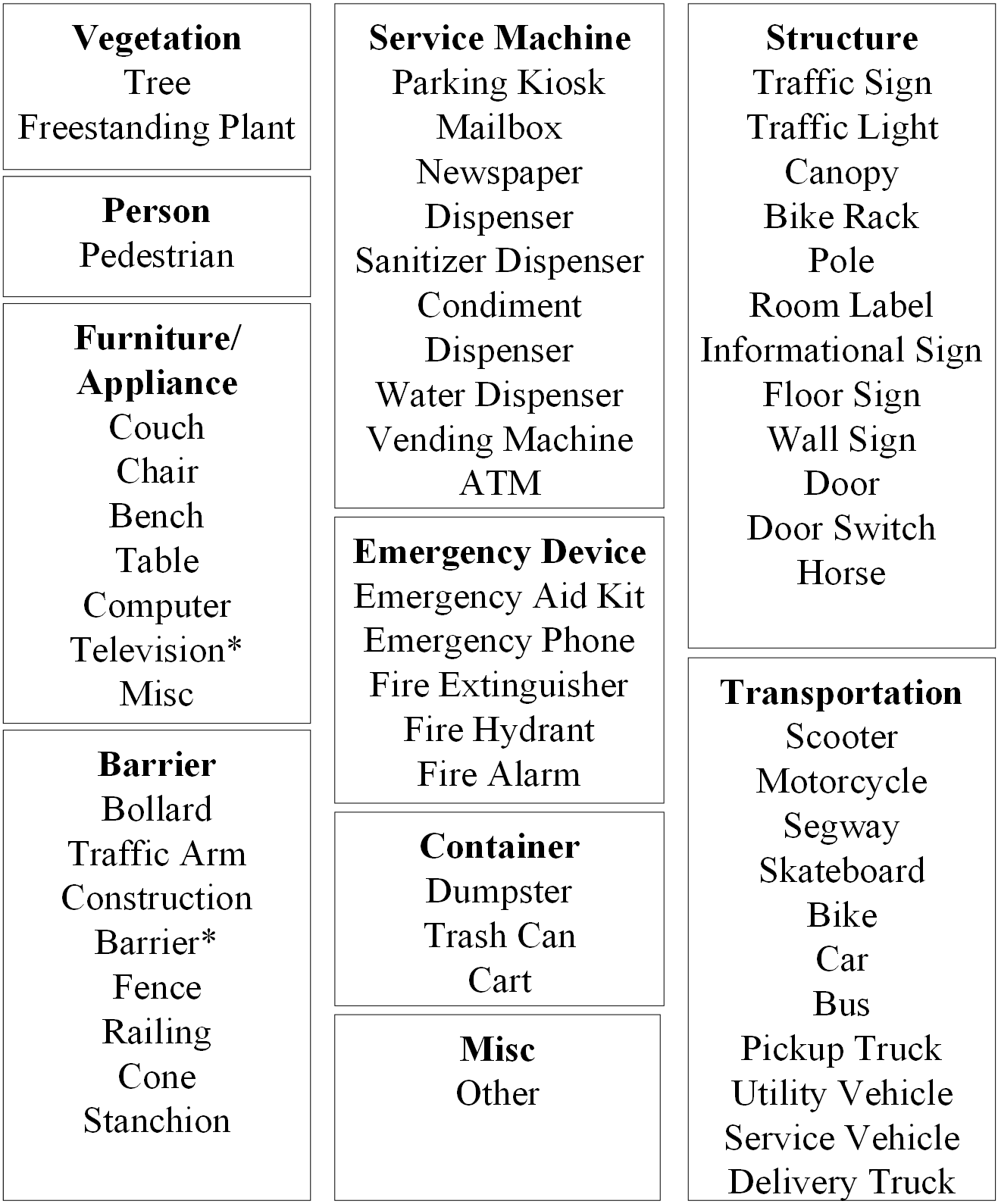}
    \caption{Full 3D bounding box object list. Bolding represents the topological category for the semantic classes below. Classes with an asterisk (*) do not have annotations in \coda{}.}
    \figlabel{fullobjectlist}
\end{figure}

\begin{figure*}[htbp]
    \centering
    \def\svgwidth{0.75\linewidth}
    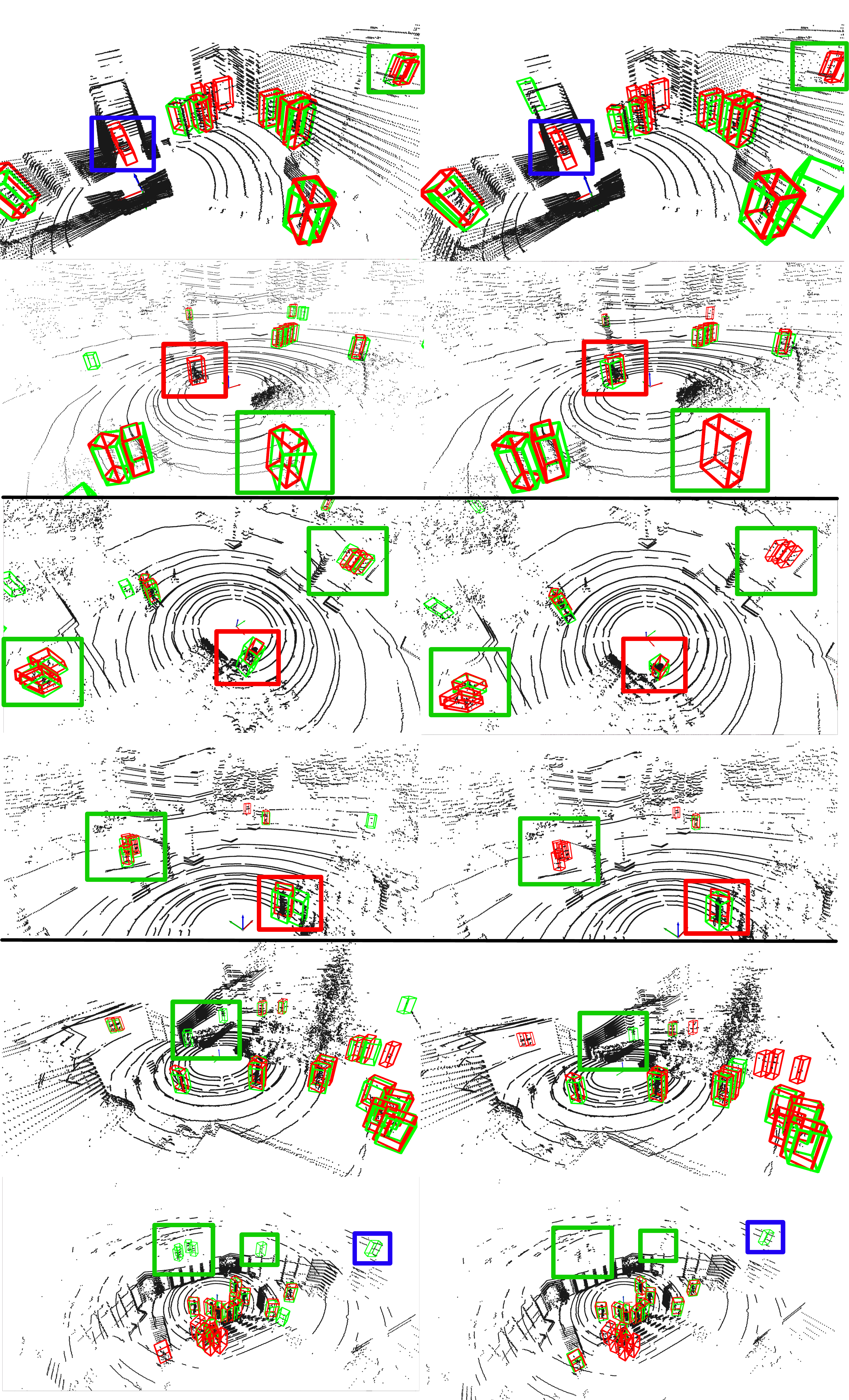    
    \caption{Qualitative 3D object detection comparison on JRDB~\cite{JRDB} after pretraining on \coda{} (left) and Waymo~\cite{Waymo} (right). We provide examples for false negative (FN), true positive (TP), and false positive (FP) and show predictions in green and ground truth annotations in red. Areas where \coda{} models perform favorably are shown in green and areas where Waymo models perform favorably are shown in red. Areas where both detectors perform similarly are shown in blue.}
    \figlabel{jrdbtpfpfn}
\end{figure*}

\subsection{Annotation Ontology}
\seclabel{app_annotation_onotology}

\coda{} is annotated with object classes from \figref{fullobjectlist} and terrain classes from \figref{fullsemseglist}. We include RGB examples of each terrain class in \figref{semantic2dexamples} by projecting the 3D point cloud annotation to the corresponding 2D image. For 2D examples of annotated objects, we refer the reader to the annotation instruction document in the data report~\cite{coda2023tdr}.

\subsection{Qualitative Object Detection Results} \seclabel{app_qualitative_results}

We supplement the quantitative cross-dataset object detection evaluation in \tabref{jrdb_hyperparameter_summary} with qualitative results to vindicate our claim that pretraining on \coda{} provides better sparse detection performance. \figref{jrdbtpfpfn} contains examples of false negative, true positive, and false positive detections from our pre-trained models on JRDB. The first two examples demonstrate that detecting non-ground plane pedestrians and pedestrians on bicycles is challenging. This is because our pretraining datasets annotate pedestrians on cyclists differently than JRDB and do not contain examples of pedestrians below the ground plane. The last four examples show that \coda{} pre-training outperforms Waymo at detecting pedestrians that are sparse in LiDAR point cloud data.

\subsection{Ground Truth Annotation Visualizations}
\seclabel{app_gt_visualizations}

We conclude with visualizations of the ground truth 3D bounding box annotations in \coda{}. \figref{coda_gt_visualizations_part1} and \figref{coda_gt_visualizations_part2} showcase images from each annotated sequence in \coda{}. We characterize the dataset and annotation diversity in \figref{locationweathercount}, \figref{distributionobjects}, \figref{objectheatmap} and \figref{objecthistogram}, \figref{semantichistogram}. The data diversity, large semantic class list, and real-world nature of \coda{} make it a comprehensive dataset and benchmark for egocentric perception algorithms.

\begin{figure*}[htbp]
\centering
\includegraphics[width=0.95\textwidth]{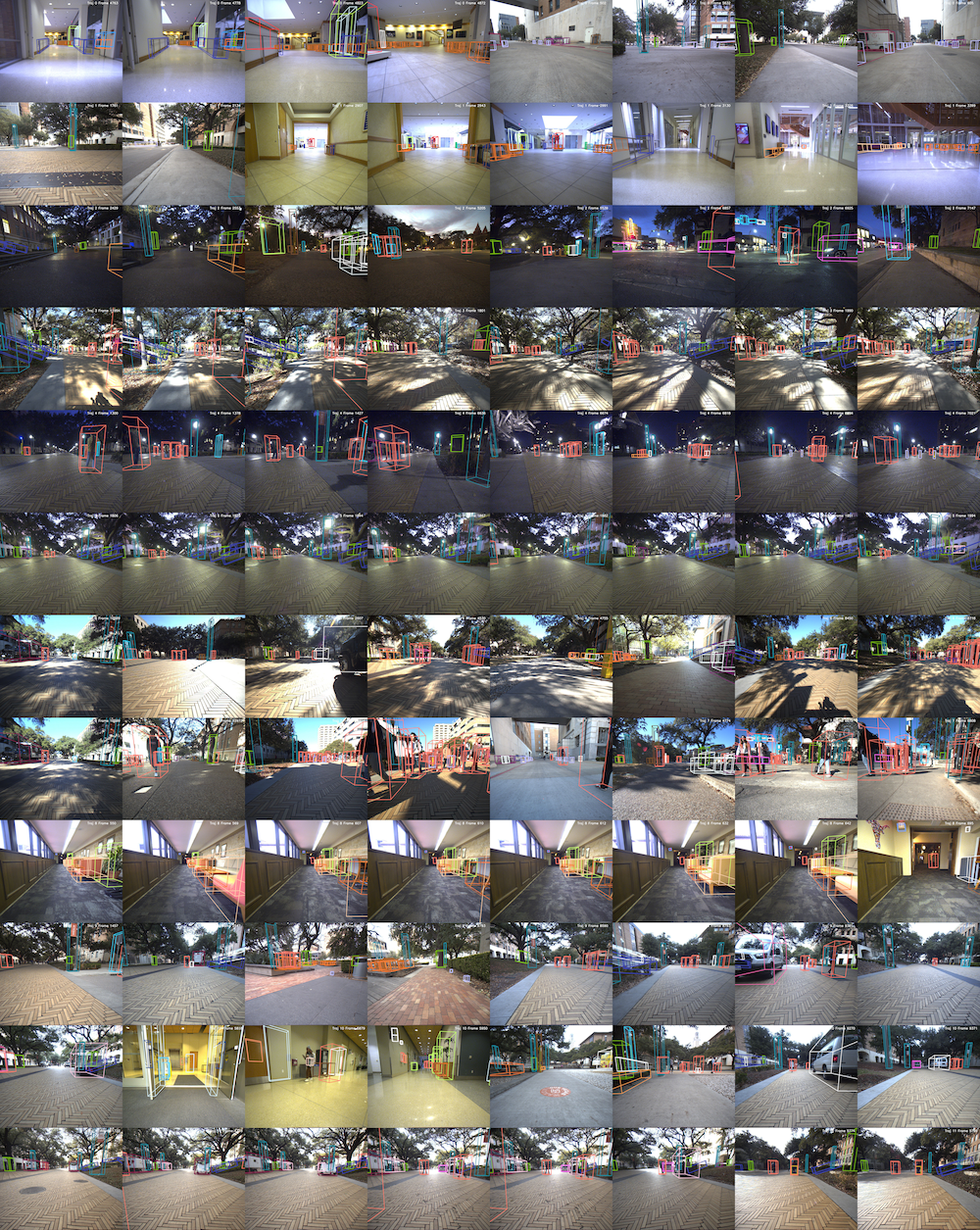}
\caption{Scenes from all annotated sequences in \coda{}. Each row contains images from the same sequence. These sequences are collected from the perspective of an urban robot in indoor and outdoor environments: cafeterias, public workspaces, sidewalks, and libraries. We quantify the dataset's weather, lighting, and viewpoint diversity in \figref{locationweathercount} and \figref{objecthistogram}. The data diversity, large semantic class list, and real-world nature of \coda{} make it a comprehensive dataset and benchmark for egocentric perception algorithms.}
\figlabel{coda_gt_visualizations_part1}
\end{figure*}

\begin{figure*}[htbp]
\centering
\includegraphics[width=0.95\textwidth]{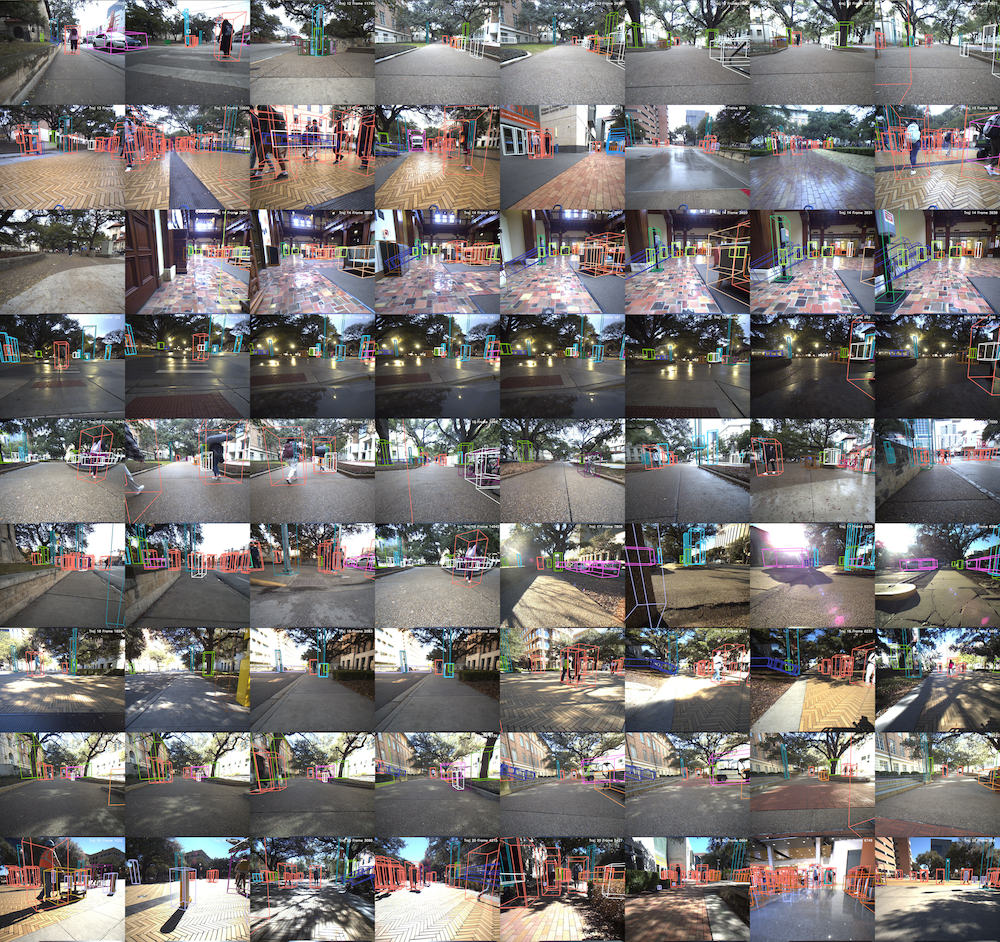}
\caption{Scenes from all annotated sequences in \coda{} (cont.)}
\figlabel{coda_gt_visualizations_part2}
\end{figure*}

\end{document}